\documentclass[lettersize,journal]{IEEEtran}
\usepackage{amsmath,amsfonts}
\usepackage{colortbl} 
\usepackage{color}
\usepackage[ruled]{algorithm2e}
\usepackage{array}

\usepackage{textcomp}
\usepackage{booktabs}
\usepackage{stfloats}
\usepackage{url}
\usepackage{verbatim}
\usepackage{caption}
\usepackage{xcolor}
\usepackage{graphicx}
\usepackage{threeparttable}
\usepackage{cite}
\usepackage{multirow}

\usepackage{subfigure}

\usepackage{fancyhdr}
\usepackage{kantlipsum} 

\fancypagestyle{arxivpolicy}{
    \fancyhf{} 
    \fancyfoot[L]{\footnotesize © 2025 IEEE. Personal use of this material is permitted. Permission from IEEE must be obtained for all other uses, in any current or future media, including reprinting/republishing this material for advertising or promotional purposes, creating new collective works, for resale or redistribution to servers or lists, or reuse of any copyrighted component of this work in other works.}
}
\begin{document}

\title{REMAC: Reference-based Martian Asymmetrical \\ Image Compression}

\author{Qing Ding, 
        Mai Xu\textsuperscript{*}, \IEEEmembership{Senior Member, IEEE}, 
        Shengxi Li, \IEEEmembership{Member, IEEE}, 
        Xin Deng, \IEEEmembership{Member, IEEE},
        Xin Zou
\IEEEcompsocitemizethanks{This work was supported by NSFC under Grants 62231002, 62522101, 62372024, 62206011, and 62450131, and Beijing Natural Science Foundation under Grant L223021, and the Fundamental Research Funds for the Central Universities.

\IEEEcompsocthanksitem \textsuperscript{*}Corresponding Author: M. Xu (MaiXu@buaa.edu.cn).
\IEEEcompsocthanksitem Q. Ding, M. Xu, S. Li, X. Deng, X. Zou are with the School of Electronic and Information Engineering, Beihang University, Beijing, 100191, China (email: qingding@buaa.edu.cn; MaiXu@buaa.edu.cn; ShengxiLi2014@gmail.com; cindydeng@buaa.edu.cn, zouxin501@163.com). X. Zou is also with the Beijing Institute of Spacecraft System Engineering,  Beijing, 100094, China.

}
}

\markboth{Journal of \LaTeX\ Class Files,~Vol.~14, No.~8, August~2021}%
{Shell \MakeLowercase{\textit{et al.}}: A Sample Article Using IEEEtran.cls for IEEE Journals}

\maketitle
\thispagestyle{arxivpolicy}
\begin{abstract}
To expedite space exploration on Mars, it is indispensable to develop an efficient Martian image compression method for transmitting images through the constrained Mars-to-Earth communication channel. Although the existing learned compression methods have achieved promising results for natural images from earth, there remain two critical issues that hinder their effectiveness for Martian image compression: 1) They overlook the highly-limited computational resources on Mars; 2) They do not utilize the strong \textit{inter-image} similarities across Martian images to advance image compression performance. Motivated by our empirical analysis of the strong \textit{intra-} and \textit{inter-image} similarities from the perspective of texture, color, and semantics, we propose a reference-based Martian asymmetrical image compression (REMAC) approach, which shifts computational complexity from the encoder to the resource-rich decoder and simultaneously improves compression performance. To leverage \textit{inter-image} similarities, we propose a reference-guided entropy module and a ref-decoder that utilize useful information from reference images, reducing redundant operations at the encoder and achieving superior compression performance. To exploit \textit{intra-image} similarities, the ref-decoder adopts a deep, multi-scale architecture with enlarged receptive field size to model long-range spatial dependencies. Additionally, we develop a latent feature recycling mechanism to further alleviate the extreme computational constraints on Mars. Experimental results show that REMAC reduces encoder complexity by 43.51\%  compared to the state-of-the-art method, while achieving a BD-PSNR gain of 0.2664 dB.

\end{abstract}

\begin{IEEEkeywords}
Mars, Mars vision tasks, lossy image compression, asymmetric autoencoder, reference-guided entropy model, reference-aware decoder.
\end{IEEEkeywords}

\section{Introduction}
\IEEEPARstart{N}{owadays}, humans are passionate about exploring Mars through data collected by landing rovers, which are transmitted to Earth for further scientific research and science popularization \cite{MarsExp1,MarsExp2,MarsExp3}. Moreover, images are the majority of the data transmitted over the Mars-to-Earth communication channel. Unfortunately, the computational resources of the planetary rovers on Mars are extremely limited due to the long distance between the two planets. To successfully transmit high-fidelity images for Mars exploration, it is vital to develop an efficient Martian image compression method that significantly reduces the encoder complexity while embracing superior compression performance.

\begin{figure}[t]
	\centering
	\includegraphics[height=2.4in]{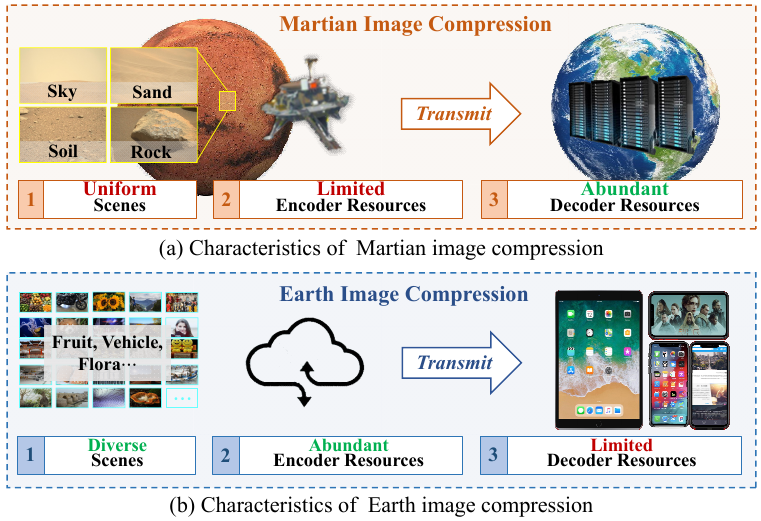}
 
	\caption{The characteristics of Martian and Earth image compression.}
	\label{fig:1}
\end{figure}

Compared to Earth images, Martian image
compression has its own distinct characteristics, which are illustrated in Fig. \textcolor{red}{\ref{fig:1}}. Specifically, the differences are primarily reflected in the following three aspects: 1) \textit{Scene characteristics}: Martian image compression typically deals with uniform-scene images, due to the monotonous and homogeneous landscapes of Mars, whereas the diverse scenes must be accommodated for Earth images. 2) \textit{Encoder resource constraints}: The Martian encoding process is severely constrained by the limited computational capability of the landing rover, while Earth image compression can leverage abundant resources through distributed cloud computing infrastructures. 3) \textit{Decoder resources constraints}: Martian image decoding benefits from the sufficient computational resources of ground station platforms, in contrast to Earth image compression that must decode under strict resource limitations when deployed in front-end devices \cite{kim2020efficient}. These differences render existing compression algorithms unsuitable for Martian images, necessitating the development of a more efficient compression approach specifically adapted to Martian image compression characteristics.

The existing traditional codecs are based on image transform, such as high-efficiency video coding (HEVC) \cite{hevc}, and versatile video coding (VVC) \cite{vvc}. However, these codecs cannot be adaptive to image content,  thus hindering
their effectiveness for Martian image compression. Taking advantage of the great success of deep learning, the learning-based compression methods \cite{vcip,cheng,wacnn,hyperprior} have been proposed, which are trained with a large number of images, thus adaptive to image content for improving the rate-distortion (R-D) performance. Due to the abundance of Martian images collected by \cite{vcip}, state-of-the-art learning-based image compression methods can be adopted to compress Martian images efficiently. Although these learning-based compression methods have made considerable progress, they overlook the differences between Earth and Martian image compression as mentioned above. This leads to two critical problems that hinder their efficiency for Martian image compression. One problem is that these methods do not fully exploit the distinct Martian characteristic of scene uniformity. The other problem is that these methods do not consider the limited computational resources of encoders and the sufficient computational resources of decoders. For practical Martian applications, lightweight encoders are thus required, but high computational complexity can be afforded for decoding the compressed Martian images.

In this paper, we propose a reference-based Martian asymmetrical image compression (REMAC) approach to reduce encoder complexity and improve compression performance. The design of REMAC is guided by the distinctive characteristics of Martian image compression as illustrated in Fig. \textcolor{red}{\ref{fig:1}}. We begin with a comprehensive data analysis of Martian images, revealing strong \textit{intra-} and \textit{inter-image} similarities in terms of texture, color, and semantics. Motivated by these observations, REMAC is designed to exploit both types of similarities to shift computational complexity from the resource-limited encoder to the resource-rich decoder, while achieving improved compression performance. Specifically, the \textit{inter-image} similarities inspire the design of a reference-guided entropy module and a ref-decoder, which enhance entropy estimation and decoding reconstruction by leveraging reference images available at both the encoder and decoder sides. By transferring useful information from reference images, the encoder avoids redundant computation on the input image, while the decoder performs additional operations on the reference image to compensate, effectively achieving computational asymmetry. Meanwhile, the \textit{intra-image} similarities motivate a deep, multi-scale architecture in the ref-decoder, enabling a larger receptive field size to capture long-range spatial dependencies within individual images. Finally, to further minimize encoder complexity during the inference stage, we propose a \emph{latent feature recycling} mechanism that repurposes the encoder's already-extracted latent features for deep reference image selection, such that specific feature extraction can be discarded during inference.

In summary, the contributions of our work are three-fold as follows.

\begin{enumerate}
	\item Motivated by the strong \emph{inter-image} similarities across Martian images, we propose a novel reference-guided entropy module and a ref-decoder that employ reference images to enhance entropy estimation and decoding reconstruction, while reducing encoder complexity.
	\item Motivated by the strong \emph{intra-image} similarities, we propose a deep, multi-scale architecture in the ref-decoder, enabling a larger receptive field size to model long-range spatial dependencies within individual images.
 	\item We propose a latent feature recycling mechanism to repurpose the encoder's latent features for deep reference selection, such that specific feature extraction can be eliminated to reduce encoder complexity during inference.

\end{enumerate}

\begin{figure*}[t]
	\centering
	\includegraphics[height=1.7in]{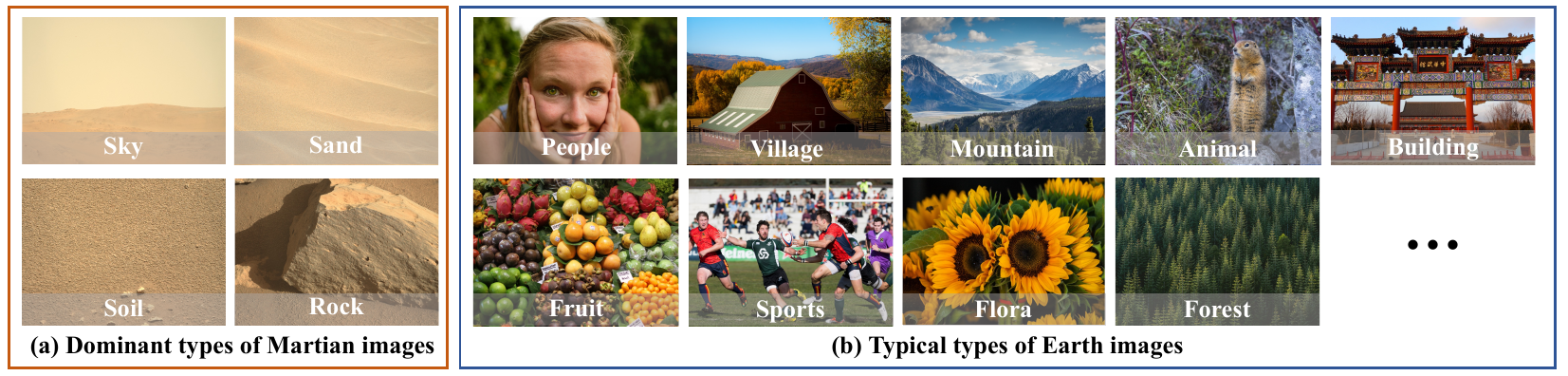}
 
	\caption{Comparison of representative Martian (left) and Earth (right) images.}
	\label{fig:earth}
\end{figure*}

\begin{figure*}[t]
	\centering
	\includegraphics[width=6.8in]{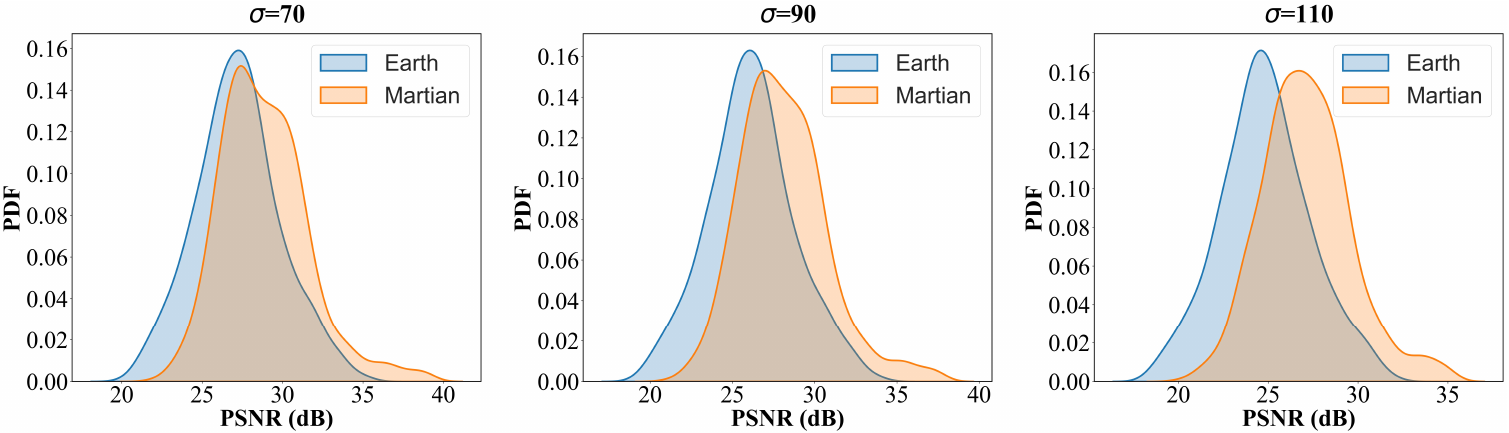}
	\caption{The distributions of  PSNR values on Earth and Martian images with Gaussian noise levels of  $\sigma=70$, $\sigma=90$, and $\sigma=110$.}
	\label{fig:2}
\end{figure*}

\begin{figure}[t]
	\centering
	\includegraphics[width=3.5in]{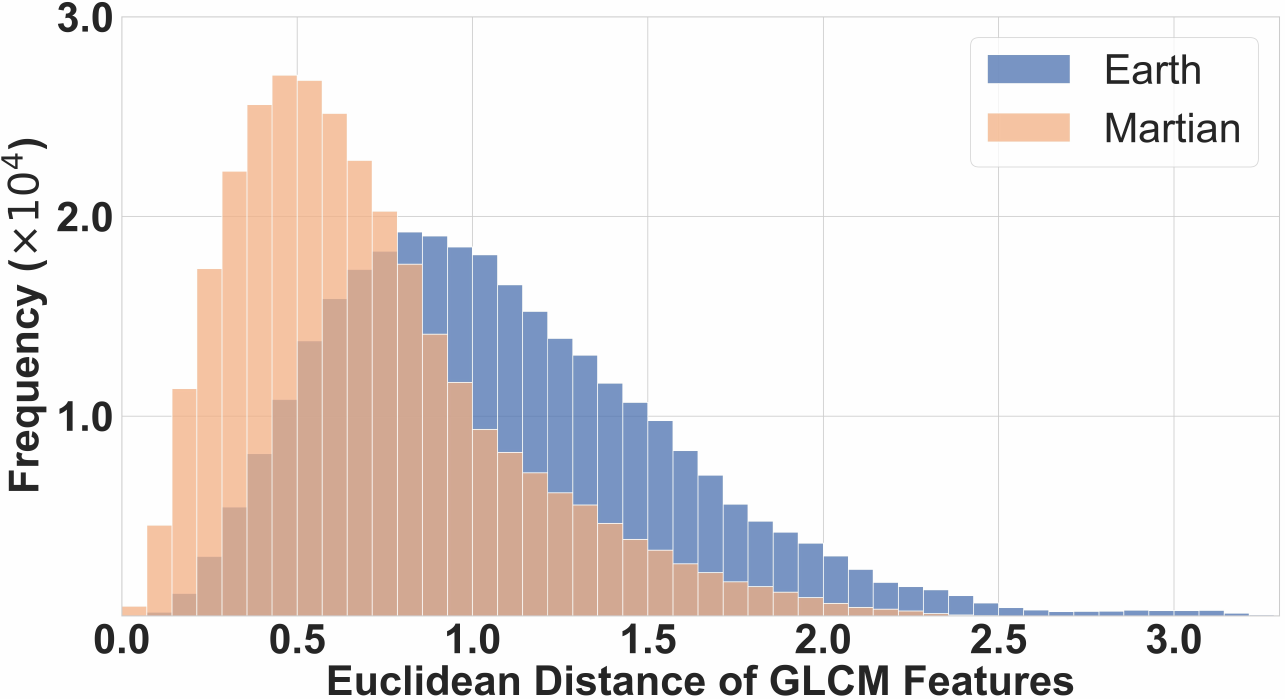}
	\caption{The histograms of Euclidean distances of GLCM features for Martian and Earth images.}
	\label{fig:glcm}
\end{figure}

\section{Related Work}
\subsection{Learning-based Image Compression}
Existing learning-based image compression (LIC) methods can be categorized into three major optimization perspectives: transform-focused, quantization-focused, and entropy-coding-focused designs. In this section, we review the transform-focused and entropy-coding-focused designs, which are most relevant to our study.

Transform-focused approaches aim to advance the transform module for improving latent representation quality, thereby enhancing compression efficiency. Ballé \emph{et al.} \cite{balle1} first proposed an end-to-end variational autoencoder framework for compression, laying the foundation for subsequent transform-focused models. Building on this, Cheng \emph{et al.} \cite{cheng} introduced attention mechanisms to enhance detail preservation during transformation. More recently, the rise of Vision Transformer \cite{ViT} has inspired methods. Zou \emph{et al.} \cite{wacnn} designed a window-based attention block to capture both local and global dependencies. Liu \emph{et al.} \cite{Liu2023} proposed a hybrid Transformer-CNN block to take advantages of  both CNN and Transformer models. Meanwhile, numerous transform-focused image compression frameworks have emerged for remote sensing images, such as using frequency separation \cite{TGRSIC_frequency}, Markov-inspired attention \cite{TGRSIC_markov} and multilevel similarity constraint \cite{TGRSIC_multilevel}. Furthermore, Ding \emph{et al.} \cite{vcip} exploited non-local attention block \cite{nln} for Martian image compression, utilizing the strong similarity among Martian image patches.

Entropy-coding-focused approaches aim for more accurate entropy estimation to enhance compression efficiency. Ballé \emph{et al.} \cite{hyperprior} introduced a hyperprior model to utilize the dependency structure in the compact latent for improving entropy estimation. Minnen \emph{et al.} \cite{autoregressive} extended this with a spatial autoregressive model to fully utilize the contextual information of the transformed latent. However, this method is very time-consuming. To address this drawback, the channel-wise \cite{channel-wise}, checkerboard \cite{checkboard}, and space-channel \cite{ELIC} autoregressive models were proposed to accelerate the process.

Although substantial progress has been made, most existing LIC methods are designed for Earth images, which exhibit high scene diversity and limited \textit{inter-image} redundancy. Consequently, these methods do not exploit the high \textit{inter-image} similarity inherent in Martian images. Moreover, many of these methods rely on computationally heavy encoders. Together, these limitations lead to suboptimal compression performance and hinder practical deployment on Mars.

\subsection{Reference-based Methods}
As lossy compression inevitably discards image details during quantization, reference images can be leveraged to recover the lost information for better compression performance. The closest area to this is reference-based image super-resolution (RefSR).

In RefSR, a high-resolution (HR) reference image is utilized to assist the super-resolution of a low-resolution (LR) image. Assuming that reference images are very similar to target HR images, Zheng \emph{et al.} \cite{CrossNet} proposed CrossNet, which estimated the flow between LR and HR images and warped the HR images in a cross-scale manner. To handle the RefSR task with less similar reference images, Zhang \emph{et al.} redefined the RefSR task as a neural texture transfer problem \cite{SRNTT}, utilizing semantically relevant textures for texture transfer. Inspired by advancements in attention mechanisms \cite{AllYouNeed}, Zhang \emph{et al.} \cite{TTSR } introduced a texture transformer block to perform adaptive transfer of relevant textures through attention maps. However, these methods relied on spatial alignment or dense patch matching to locate the correspondence between reference and LR images, which was time-consuming. To accelerate this process, Lu \emph{et al.} \cite{MASA} proposed the MASA-SR method, which conducted patch matching in a coarse-to-fine manner. Despite their effectiveness in image enhancement, these RefSR methods are trained solely for reconstruction quality and do not consider the compression artifacts caused by the encoder. When directly applied to image compression, they result in a decoupled two-stage compression pipeline where the encoder and decoder are optimized independently, lacking end-to-end rate-distortion optimization and leading to suboptimal compression performance.

Currently, only two works \cite{li2023rfd,underwater} have explored reference-based image compression, both specifically designed for underwater images. These methods rely heavily on underwater physical priors (e.g., using the underwater scattering model), focus primarily on compression efficiency, and overlook the computational costs, making them fundamentally unsuitable for Martian images. There are three critical limitations when applying them to Martian images: 1) They employ patch-level reference selection, causing high computational costs and failing to capture global contextual modeling. 2) They adopt encoder-heavy architectures with a complex encoder and a lightweight decoder, making them unsuitable for the constrained encoder resources on Mars. 3) They only partially integrate reference information: one \cite{li2023rfd} applies it only to the transform module, the other \cite{underwater} only to the entropy coding module, failing to exploit the benefits of joint integration.

Motivated by the strong \textit{inter-image} similarity across Martian images, we propose REMAC, a reference-based image compression approach specifically designed for Martian images. REMAC introduces a jointly optimized architecture that addresses all three limitations of prior works. 1) It employs image-level reference selection to save computational costs and enable global contextual modeling. 2) It adopts a decoder-heavy design to shift computational costs from the resource-limited encoder (on Mars) to the resource-rich decoder (on Earth), suitable for practical deployment on Mars. 3) It fully integrates reference information into both the entropy estimation and the decoding reconstruction, enabling efficient exploitation of reference images for image compression. As a result, REMAC achieves efficient image compression while respecting the practical constraints of Martian image compression, representing a significant advancement over existing reference-based image compression methods.

\subsection{Lightweight Learning-based Image Compression}

To enable practical deployment of LIC models under resource constraints, numerous lightweight LIC methods have been proposed. Existing lightweight compression methods can be broadly classified into two categories: model compression techniques and efficient architecture designs.

Model compression techniques adopt network pruning and weight quantization to reduce model parameters and complexity. To adapt LIC models to resource-constrained front-end devices, Kim \textit{et al.} \cite{kim2020efficient} used weight and filter pruning to reduce its decoder. Observing that the hyperprior path in the LIC model is over-parametrized, Luo \textit{et al.} \cite{luo2022memory} proposed to prune it for lower model parameters. To further achieve optimal pruning results in the hyperprior path, Luo \textit{et al.} \cite{luo2023pts} proposed to search for different appropriate pruning thresholds for each bpp-point LIC model. To further improve the deployment efficiency of LIC models, Fang \textit{et al.} \cite{fang2023fully} proposed an effective quantization method for weights and activations. 

Efficient architecture designs employ specialized convolution operations and network architectures to make the LIC models lightweight. Inspired by efficient design in the self-attention module, He \textit{et al.} \cite{he2023efficient} incorporated the efficient attention module into transformer-based LIC models. This significantly reduces the high computational costs of the self-attention mechanism. To reduce the decoder complexity, Fu \textit{et al.} \cite{fu2023asymmetric} and Wang \textit{et al.} \cite{wang2024asymllic} proposed asymmetrical LIC models, in which the decoder network contains shallower layers than the encoder network. More recently, Bao \textit{et al.} \cite{bao2025shiftlic} adopted parameter-free shift operations to replace large-kernel convolutions, thereby reducing the computational costs.

While these methods are effective in general scenarios, they lack task specificity, particularly in domain-specific task like Martian image compression. These generic designs exhibit two major limitations for Martian image compression: (1) they fail to specifically exploit image characteristics, such as the high \textit{inter-image} similarity across Martian images, for compression efficiency, and (2) most employ symmetrical architectures, and even the few asymmetrical methods only focus on reducing decoder complexity, limiting their suitability for Martian compression scenarios where the encoder is resource-constrained. These limitations motivate our scenario-driven design, which explicitly utilizes Martian image characteristics and abundant decoder resources to make the encoder lightweight while maintaining compression performance.

\section{Data Analysis}
In this section, we thoroughly analyze the characteristics of Martian images, looking for the potential to improve compression efficiency and reduce encoder complexity. Specifically, we select 784 Earth images from the DIV2K \cite{div2k} dataset, ensuring their width and height are no less than the resolution of the Martian images in the MIC \cite{vcip} dataset. Meanwhile, we randomly select 784 Martian images from the MIC \cite{vcip} dataset for analysis. For fair comparisons, all selected Earth images are cropped to 1600 $\times$ 1152 pixels, matching the resolution of Martian images. To facilitate intuitive comparison between the two planets, representative Martian and Earth images are presented in Fig. \textcolor{red}{\ref{fig:earth}} to highlight their visual distinctions. As shown, Martian images exhibit higher color concentration and are dominated by four types: sky, sand, soil, and rock. In contrast, Earth images display significantly greater diversity in both color distribution and scene categories. Building upon these visual observations, our subsequent quantitative analysis reveals three important findings.

\textbf{Finding 1:} Compared to Earth images, Martian images exhibit stronger \emph{intra-} and \emph{inter-image} texture similarity.

\textbf{Analysis:} To measure the \emph{intra-image} texture similarity, we apply the block-matching and 3-dimensional filtering (BM3D) algorithm \cite{BM3D} on both noisy Earth and Martian images. The BM3D algorithm can efficiently remove noise from image that posses more similar patches. Therefore, the higher denoising performance indicates  the stronger  \emph{intra-image} texture similarity. We apply multiple levels of additive Gaussian noise to both Earth and Martian images and subsequently utilize the  BM3D algorithm to remove  noise. The denoising performance is evaluated using the Peak Signal-to-Noise Ratio (PSNR) metric. Fig. \textcolor{red}{\ref{fig:2}} shows the distributions of  PSNR values on Earth and Martian images with Gaussian noise levels of  $\sigma=70$, $\sigma=90$, and $\sigma=110$. As can be seen, across multiple noise levels, the BM3D algorithm achieves higher PSNR values on Martian images than on Earth images, indicating that Martian images possess stronger \emph{intra-image} similarity than Earth ones.

To further assess \emph{inter-image} texture similarity, we use the gray-level co-occurrence matrix (GLCM) \cite{GLCM} to extract six texture features, including contrast, dissimilarity, homogeneity, energy, correlation, and angular second moment. Subsequently, these features are normalized and concatenated into a vector, which is used to represent the image texture. Then, we calculate the Euclidean distance between these vectors captured from different images. A smaller distance between two images indicates a higher degree of \emph{inter-image} texture similarity. We also visualize the distribution of these distances in Fig. \textcolor{red}{\ref{fig:glcm}}. As depicted, the distribution of distances for Martian images lies to the left of that for Earth images, indicating that most distances are smaller. This confirms that Martian images possess stronger \emph{inter-image} texture similarity compared to Earth ones.

\begin{figure*}[t]
	\centering
	\includegraphics[width=6.8in]{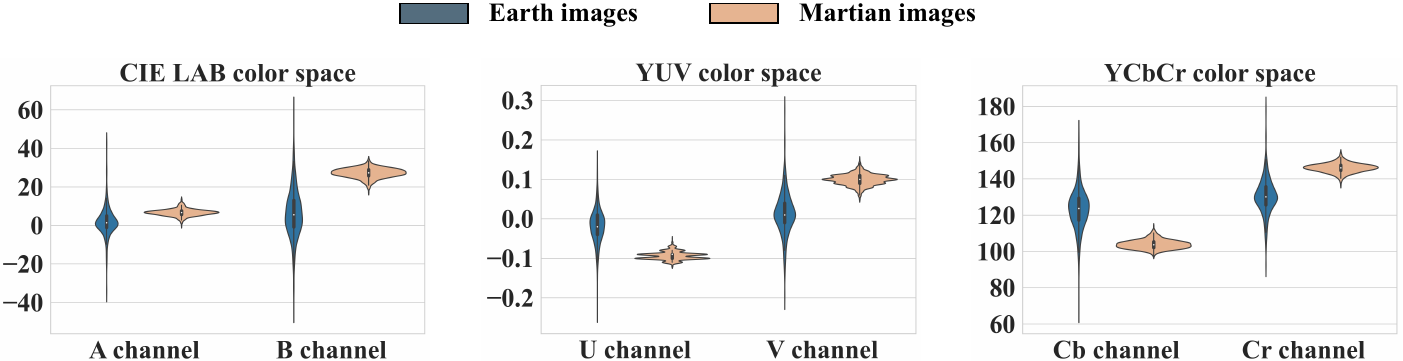}
	\caption{The violin charts of mean values in color channels across different color spaces.}
	\label{fig:4}
\end{figure*}

\textbf{Finding 2:} Compared to Earth images, Martian images display stronger \emph{intra-} and \emph{inter-image} color similarity.

\textbf{Analysis:} To analyze the color similarity between images, we compare the color distributions of Earth and Mars images in the CIE LAB, YUV, and YCbCr color spaces. Specifically, we first convert the RGB images into these color spaces, to decompose the image data into luminance and chrominance components. Since the chrominance components represent most color information, we calculate the standard deviation (STD) for each image across the chrominance channels to evaluate the \emph{intra-image} color similarity. The results are presented in Tab. \textcolor{red}{\ref{tab:1}}. As seen in this table, the Martian images have lower mean STD values in each color space, demonstrating that the Martian images have a higher degree of \emph{intra-image} color similarity. Furthermore,  to compare \emph{inter-image} color similarity, we calculate the mean values of the chrominance channels for each image. The corresponding violin charts of mean values for Earth and Martian images are plotted in Fig. \textcolor{red}{\ref{fig:4}}. As illustrated, Martian images display more concentrated mean values in chrominance channels in each color space, indicating stronger \emph{inter-image} color similarity among Martian images.

\begin{table}[tp]
\begin{center}

\caption{The average STD value across two chrominance channels in each color space.}
\label{tab:1}
{
\setlength{\tabcolsep}{8pt} 
\renewcommand{\arraystretch}{1.3} 
\small
\begin{tabular}{|c|c|c|c|}
\cline{1-4}
\multicolumn{1}{|c|}{Color Space} &\multicolumn{1}{c|}{Chrominance Channel} &
\multicolumn{1}{c|}{Martian} &
\multicolumn{1}{c|}{Earth} \\
\hline
\multirow{2}{*}{CIE LAB} & A & 3.0024 & 8.4402 \\ 
                         & B & 4.5352 & 13.9493 \\
                         \hline
\multirow{2}{*}{YUV}     & U & 0.0171 & 0.0478 \\
                         & V & 0.0235 & 0.0632 \\
                         \hline
\multirow{2}{*}{YCbCr}   & Cb & 4.4179 & 12.2660 \\
                         & Cr & 4.2587 & 11.4953 \\
\hline
\end{tabular}}
\end{center}

\end{table}

\begin{figure}[t]
	\centering
	\includegraphics[width=3.5in]{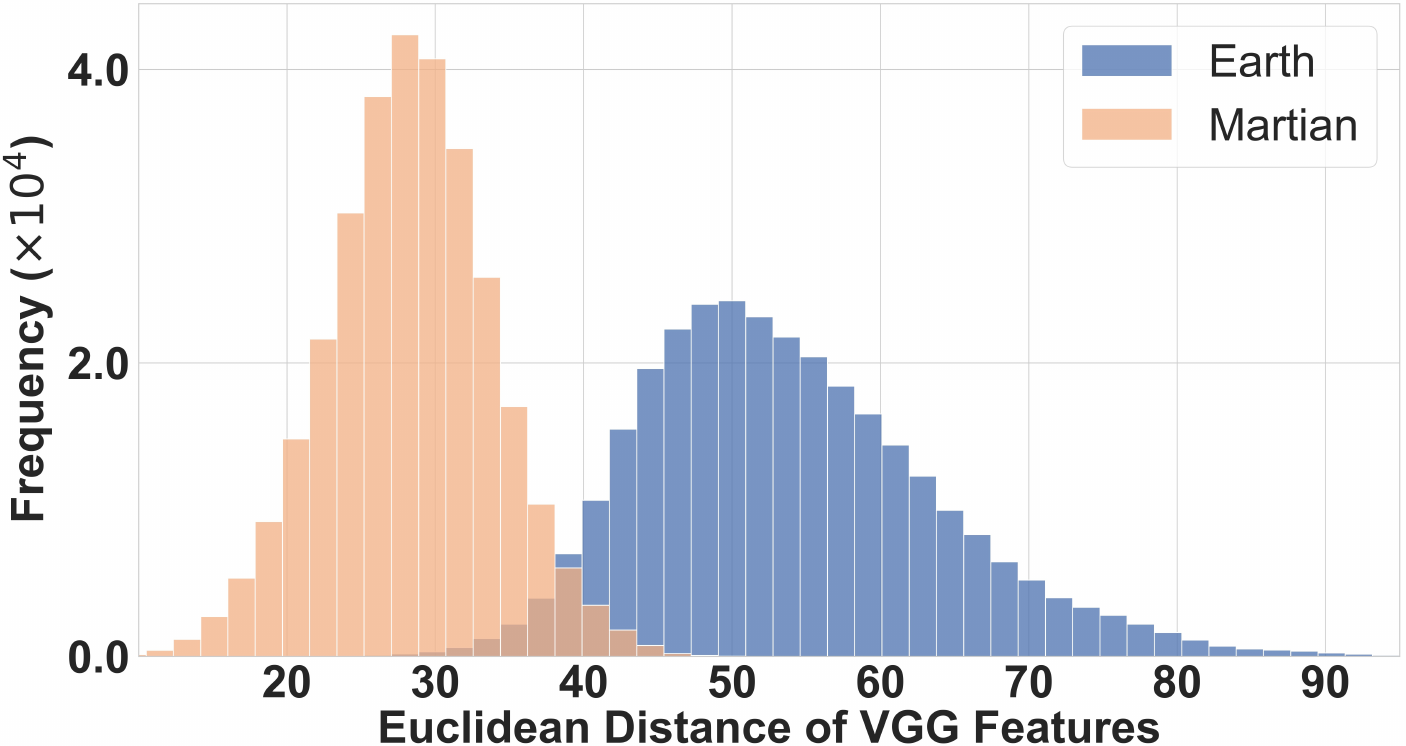}
 
	\caption{The histograms of Euclidean distances of VGG features for Martian and Earth images.}
	\label{fig:vgg}
\end{figure}

\begin{figure*}[t]
	\centering
	\includegraphics[width=7in]{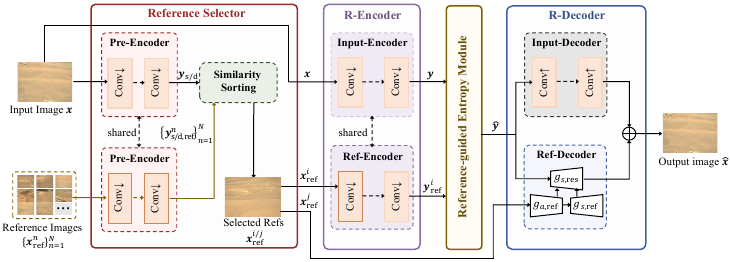}
 
	\caption{Framework of the proposed REMAC. Here, \textit{R-Encoder} and \textit{R-Decoder} denote the proposed reference-aware encoder and decoder, respectively. Each ``Conv$\downarrow$'' consists of a down-sampling convolutional layer followed by a GDN layer, except for the last one, which contains only the convolutional layer. Similarly, each ``Conv$\uparrow$'' comprises an up-sampling convolutional layer followed by an inverse GDN layer, except for the last one.}
	\label{fig:framework}
\end{figure*}

\textbf{Finding 3:} Compared to Earth images, Martian images display stronger semantic similarity.

\textbf{Analysis:} To analyze the semantic similarity of Martian images, we employ the well-known VGG16 classification model \cite{VGG} to extract semantic features from each image. The Euclidean distance between these features is then computed to measure the degree of semantic similarity. The smaller Euclidean distance indicates stronger  semantic similarity. Specifically, we utilize the features from the last ReLU layer of the VGG16 model to extract the semantic features.  The distributions of the Euclidean distances for Martian and Earth images are shown in Fig. \textcolor{red}{\ref{fig:vgg}}. As can be seen, the distances between Martian images  are far smaller than those between Earth images, indicating that the Martian images have stronger semantic similarity.

\section{The Proposed Approach}
Based on the above findings, we propose the REMAC approach to fully utilize the \emph{intra-} and \emph{inter-image} similarities. This approach leverages reference images that share similarities with the input images to fully exploit the above Martian characteristics, enabling efficient image compression with a lightweight encoder. In Section IV-A, we introduce the overall architecture of the REMAC approach. We then introduce the proposed reference selector module, reference-aware encoder (R-encoder) and decoder (R-decoder), and reference-guided entropy module in Sections IV-B, IV-C, and IV-D, respectively. Section IV-E introduces the proposed latent feature recycling mechanism and the corresponding loss functions for rate-distortion optimization with a lightweight encoder.

\subsection{Overall Architecture}
We illustrate the overall architecture of the proposed REMAC in Fig. \textcolor{red}{\ref{fig:framework}}, which consists of the reference selector, R-encoder, reference-guided entropy module, and R-decoder. More specifically, regarding the reference selector, $N$ images are randomly selected from the training set of the MIC dataset \cite{vcip} to construct a reference dataset $\{\boldsymbol{x}_{\text{ref}}^{n}\}_{n=1}^{N}$. To efficiently utilize the reference images, input-reference triplets are generated by calculating and sorting the similarity between input and reference features, which are extracted by a pre-trained encoder. Based on these triplets, one selected reference image $\boldsymbol{x}_{\text{ref}}^{i}$ and the corresponding input image $\boldsymbol{x}$ are fed into the R-encoder to obtain representative reference and input features for compression. These features are then processed by the proposed reference-guided entropy module, which improves the entropy estimation of the input features $\boldsymbol{y}$, with the guidance from the reference features $\boldsymbol{y}_{\text{ref}}^{i}$. Then, the other selected reference image $\boldsymbol{x}_{\text{ref}}^{j}$ is input to the proposed R-decoder, enhancing the reconstruction of the quantized features $\hat{\boldsymbol{y}}$. To further minimize encoder complexity during inference, we propose a latent feature recycling mechanism whereby the REMAC model is fine-tuned to directly re-purpose its latent features $\boldsymbol{y}$ and $\boldsymbol{y}_{\text{ref}}^{i}$ for deep reference selection, thus eliminating the need for deep-layer feature extraction in the reference selector.

\subsection{Reference Selector}
In the proposed reference selector, two reference images are selected to serve distinct roles: 1) one based on shallow texture features to guide decoding reconstruction, and 2) another based on deep semantic features to enhance entropy estimation. To achieve this, we employ a pre-trained encoder to extract shallow and deep features from both the input and candidate reference images. For each candidate, we compute similarity and sort them. The reference images with the highest similarity are then selected as the corresponding reference, yielding the initial triplets of input and reference images for training REMAC.

More specifically, we construct a base compression model using the hyperprior compression structure \cite{hyperprior}, with all convolutional kernels set to 3, combined with a channel-wise entropy model \cite{channel-wise}. This model is first trained on the MIC dataset \cite{vcip}, and its pre-trained encoder is then adopted as our pre-encoder, as shown in Fig. \textcolor{red}{\ref{fig:framework}}. Given an input image $\boldsymbol{x}$ and the reference dataset $\{\boldsymbol{x}_{\text{ref}}^{n}\}_{n=1}^{N}$, we feed them into the pre-encoder to obtain the shallow input features $\boldsymbol{y}_{\text{s}}$, deep input features $\boldsymbol{y}_{\text{d}}$, and shallow reference features $\{\boldsymbol{y}_{\text{s, ref}}^{n}\}_{n=1}^{N}$ and deep reference features $\{\boldsymbol{y}_{\text{d, ref}}^{n}\}_{n=1}^{N}$. For each candidate reference, we compute the similarity independently for shallow and deep features, sort them in descending order, and select the top-ranked reference images, obtain their index $j$ and $i$, forming the initial triplets of input and reference images $(\boldsymbol{x}, \boldsymbol{x}_{\text{ref}}^{{j}}, \boldsymbol{x}_{\text{ref}}^{{i}})$.

The similarity between input and reference features is measured using L1 and L2 distances:
\begin{equation}
s_{\text{s}}^{n} = \| E_\text{pre}^{\text{s}}(\boldsymbol{x})- E_\text{pre}^{\text{s}}(\boldsymbol{x}_{\text{ref}}^{n})\|_{2} = \| \boldsymbol{y}_{\text{s}}-\boldsymbol{y}_{\text{s,ref}}^{n}\|_{2},
\end{equation}
\begin{equation}
s_{\text{d}}^{n} = \| E_\text{pre}^{\text{d}}(\boldsymbol{x})- E_\text{pre}^{\text{d}}(\boldsymbol{x}_{\text{ref}}^{n})\|_{1} = \| \boldsymbol{y}_{\text{d}}- \boldsymbol{y}_{\text{d,ref}}^{n}\|_{1},
\end{equation}
where $s_{\text{s}}^{n}$ and $s_{\text{d}}^{n}$ denote the similarity scores between the shallow and deep features of the input image and the $n$-th reference image, also indicating L2 and L1 distances, respectively. The functions $E_\text{pre}^{\text{s}}(\cdot)$ and $E_\text{pre}^{\text{d}}(\cdot)$ correspond to the shallow and deep layers of the pre-encoder. Lower values indicate higher similarity, and the reference images with the minimum scores/distances are selected as the corresponding references.

\begin{figure*}[t]
	\centering
	\includegraphics[width=7.2in]{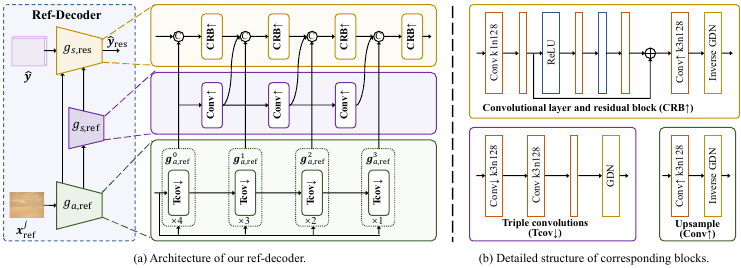}
 
	\caption{Architecture of our ref-decoder and structure of corresponding blocks. Here, \textit{k} means the kernel size, \textit{n} means the number of channels and \textit{C} means the concatenation operation. The undescribed rectangle shares an identical structure with the previous color rectangle.}
	\label{fig:7}
\end{figure*}

\subsection{Reference-aware Encoder and Decoder}
In the proposed R-encoder, we utilize the input- and ref-encoders to generate the representative features of the input and selected reference images for the subsequent entropy estimation and decoding reconstruction. Both encoders are proposed to extract the representative features and therefore, share the same architecture and weights. They consist of 4 downsampling convolutional layers followed by GDN layers. To save computational costs on the encoder side, the kernel size of each convolutional layer is set to 3. This process can be written as:
\begin{equation}
\label{featureMatching3}
\boldsymbol{y} = E(\boldsymbol{x}),
\end{equation}
\begin{equation}
\label{featureMatching4}
\boldsymbol{y}_{\text{ref}}^{\text{i}} = E(\boldsymbol{x}_{\text{ref}}^{\text{i}}),
\end{equation}
where $E$ corresponds to the functions of input and ref-encoders, which share the same weights. Meanwhile, $\boldsymbol{y}$ and $\boldsymbol{y}_{\text{ref}}^{i}$ are the generated features for compression.

In our R-decoder, we exploit the strong \emph{intra-} and \emph{inter-image} similarities in Martian images through the input- and ref-decoders, which jointly enlarge the receptive field size and integrate useful external information from reference images for superior reconstruction. The architecture of the R-decoder is shown in Fig. \textcolor{red}{\ref{fig:framework}}, in which the ref-decoder reconstructs a residual, acting as a complement of the output of the input-decoder that only utilizes the input information. The input-decoder is designed to reconstruct the compressed image $\hat{\boldsymbol{x}}$ solely from the compressed input representative features $\hat{\boldsymbol{y}}$. As the counterpart of input-encoder, the input-decoder contains 4 upsampling convolutional layers combined with inverse GDN layers. The kernel size of these convolutions is set to 3. 

To enhance reconstruction quality, the ref-decoder is proposed to extract valuable features from reference images and fuse them with the compressed input features to provide useful information. More specifically, it leverages valuable reference information in a multi-scale manner while simultaneously expanding the receptive field size to improve performance. Moreover, the detailed architecture of the ref-decoder is depicted in  Fig. \textcolor{red}{\ref{fig:7}} (a). From this figure, the ref-decoder is composed of a residual synthesis network $g_{s,\text{res}}$, a reference synthesis network $g_{s,\text{ref}}$, and multiple multi-scale reference analysis networks $g_{a,\text{ref}}^{p}$, where $p$ ranges from 0 to 3.

To leverage multi-scale reference information, we first perform multi-scale feature extraction on the selected reference image $\boldsymbol{x}_{\text{ref}}^{j}$, using multiple multi-scale reference analysis networks $g_{a,\text{ref}}^{p}$ and the reference synthesis network $g_{s,\text{ref}}$. Aiming at expanding the receptive field size for more effective feature extraction, we employ triple convolutional layers with a GDN layer, named as ``Tcov'', as the base block in multiple reference analysis networks $g_{a,\text{ref}}^{p}$, as illustrated in Fig. \textcolor{red}{\ref{fig:7}} (b). Meanwhile, we also improve the feature extraction with the assistance of the reference synthesis network $g_{s,\text{ref}}$.
\begin{figure*}[t]
	\centering
	\includegraphics[width=7.1in]{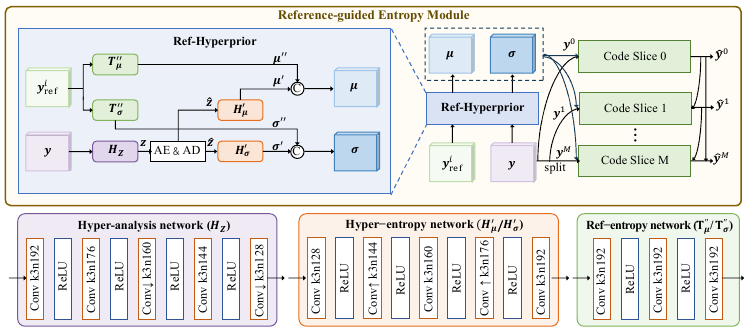}
 
	\caption{Architecture of our reference-guided entropy module. Here, \textit{k} means the kernel size, \textit{n} means the number of channels and \textit{C} means the concatenation operation.}
	\label{fig:6}
\end{figure*}

Given the extracted multi-scale reference features from the above networks, we fuse them with the compressed features $\hat{\boldsymbol{y}}$ using the residual synthesis network $g_{s,\text{res}}$ as shown in Fig. \textcolor{red}{\ref{fig:7}} (a). In this network, we feed these features into concatenation layers and adopt two convolutional layers combined with a residual block, named as ``CRB$\uparrow$'', for further fusion, as illustrated in Fig. \textcolor{red}{\ref{fig:7}} (b). Finally, we obtain the output of the ref-decoder $\hat{\boldsymbol{x}}_\text{res}$, as the complement of the input-decoder.

\subsection{Reference-guided Entropy Module}
Given the strong \emph{inter-image} similarity among Martian images, we propose the reference-guided entropy module to fully utilize the reference images, so as to provide external and valuable information for more accurate entropy estimation. Based on the initial triplets, we leverage the selected reference images to assist in predicting entropy parameters. The architecture of the reference-guided entropy module is illustrated in Fig. \textcolor{red}{\ref{fig:6}}. The input features $\boldsymbol{y}$ and selected reference features $\boldsymbol{y}_{\text{ref}}^{i}$ are fed into the ref-hyperprior to generate the entropy parameters $(\mu, \sigma)$. The newly generated $(\mu, \sigma)$ are fed into each code slice for slice-wise entropy coding.

The channel-wise autoregressive entropy model proposed in \cite{channel-wise} is a widely used learned entropy model, building upon the hyperprior entropy model \cite{hyperprior}. Compared to the hyperprior model, the channel-wise model adds slice-wise entropy coding. More specifically, it predicts the entropy parameters $(\mu, \sigma)$ of the representative features $\boldsymbol{y}$, by transmitting the side information $\boldsymbol{z}$ to the decoder side. The representative features $\boldsymbol{y}$ are then divided into $M$ slices. To perform slice-wise entropy coding, it employs the predicted $(\mu, \sigma)$ to estimate the entropy parameters of each slice. Furthermore, these parameters are conditioned on the previously decoded slices, allowing the model to utilize channel-wise contextual information in $\boldsymbol{y}$ and then achieve efficient entropy estimation.

However, this model is not effective for Martian image compression as it does not utilize the reference features $\boldsymbol{y}_{\text{ref}}^{i}$, which share similarities with the input features $\boldsymbol{y}$. Therefore, we improve it and propose the ref-hyperprior. In other words, we predict the entropy parameters $(\mu, \sigma)$ with the assistance of $\boldsymbol{y}_{\text{ref}}^{i}$, instead of solely using side information $\boldsymbol{z}$. Since the reference images can provide useful external information, this process can save coding bits with better estimation accuracy. The input and reference features are utilized to generate their respective entropy parameters, which are then concatenated to produce the updated and more accurate entropy parameters $(\mu, \sigma)$:
\begin{align}
\mu &= [\mu^{'}, \mu^{''}] = [H_{\mu}^{'}(H_{z}(\boldsymbol{y})), T_{\mu}^{''}(\boldsymbol{y}_{\text{ref}}^{i})] \notag \\
   &= [H_{\mu}^{'}(\hat{\boldsymbol{z}}), T_{\mu}^{''}(\boldsymbol{y}_{\text{ref}}^{i})], \\
\sigma &= [\sigma^{'}, \sigma^{''}] = [H_{\sigma}^{'}(H_{z}(\boldsymbol{y})), T_{\sigma}^{''}(\boldsymbol{y}_{\text{ref}}^{i})] \notag \\
   &= [H_{\sigma}^{'}(\hat{\boldsymbol{z}}), T_{\sigma}^{''}(\boldsymbol{y}_{\text{ref}}^{i})].
\end{align}
Here, $[\cdot]$ denotes the concatenation operation, $H_{z}(\cdot)$ represents the hyper-analysis network that generates the hyperprior $\boldsymbol{z}$, ${H_{\mu}^{'}}(\cdot)$ and ${H_{\sigma}^{'}}(\cdot)$ indicate the hyper-entropy networks that leverage the hyperprior $\boldsymbol{z}$ to derive the entropy parameters, and $T_{\mu}^{''}(\cdot)$ and $T_{\sigma}^{''}(\cdot)$ are the ref-entropy networks that utilize reference features to predict the entropy parameters. To reduce encoder complexity, all convolutional layers in the above networks share the same kernel size, which is 3. The detailed parameters of these networks are presented in Fig. \textcolor{red}{\ref{fig:6}}.

\subsection{Latent Feature Recycling and Optimization}
To further address the extreme limitations of encoder resources on Mars, we propose a latent feature recycling mechanism that eliminates redundant deep feature extraction during inference by re-purposing pre-computed deep input features for deep reference selection. In the pre-training stage, we first train REMAC using the above process shown as Fig. \textcolor{red}{\ref{fig:framework}}, thereby obtaining an optimized pre-trained REMAC model. Then, on the encoder side, we store the deep reference features $\{\boldsymbol{y}_{\text{ref}}^{n}\}_{n=1}^{N}$, along with the reference-based entropy parameters $\{(\mu^{''}_{n}, \sigma^{''}_{n})\}_{n=1}^{N}$. During the inference of the pre-trained REMAC model, the input image $\boldsymbol{x}$ is fed into the pre-encoder to extract representative features $\boldsymbol{y}_\text{d}$ for deep reference selection. To further reduce encoder complexity, instead of re-extracting $\boldsymbol{y}_{\text{d}}$, we reuse the latent features $\boldsymbol{y}$ and reference features $\{\boldsymbol{y}_{\text{ref}}^{n}\}_{n=1}^{N}$ captured by the pre-trained input- and ref-encoders for deep reference selection, thereby discarding the feature extraction by the deep layers of pre-encoder. This process indicates that the newly obtained deep reference image $x_{\text{ref}}^{q}$ used during inference is selected based on features extracted by the pre-trained input- and ref-encoder, different from the initial deep reference image $x_{\text{ref}}^{i}$ used in the pre-training stage. To ensure the consistency between training and inference, we freeze the pre-trained input- and ref-encoders and then fine-tune the pre-trained REMAC model using the new input-reference triplets $(\boldsymbol{x}, \boldsymbol{x}_{\text{ref}}^{{j}}, \boldsymbol{x}_{\text{ref}}^{{q}})$. Consequently, encoder complexity is reduced, while the compression performance decreases slightly. The details are illustrated in Algorithm \textcolor{red}{\ref{alg:freezing}}.

More specifically, we utilize the input- and ref-encoders, with the pre-trained parameters $\theta_{\text{E}}$, to generate latent input and reference features. First, we feed the input image $\boldsymbol{x}^{m}$ into the input-encoder to obtain latent input features $\boldsymbol{y}^{m}$. We then compute similarity scores $\boldsymbol{s}_{\text{d}}^{m}$ between $\boldsymbol{y}^{m}$ and the set of reference latent features $\{\boldsymbol{y}_{\text{ref}}^{n}\}_{n=1}^{N}$. Next, we select the reference index ${q}^{m}$ corresponding to the highest similarity (i.e., the lowest distance-based score). Furthermore, we construct new input-reference triplets $\{(\boldsymbol{x}^{m}, \boldsymbol{x}_{\text{ref}}^{j^{m}},\boldsymbol{x}_{\text{ref}}^{q^{m}})\}_{m=1}^{M}$, where $\boldsymbol{x}_{\text{ref}}^{j^{m}}$  denotes a pre-selected shallow reference image from the above pre-training stage, that remains fixed during this fine-tuning stage. Using these new triplets, we freeze the parameters of input- and ref-encoders ($\theta_{\text{E}}$), and then fine-tune the parameters of reference-guided entropy module and R-decoder ($\phi$) to obtain the final REMAC model.

For the pre-training and fine-tuning procedures, we use the commonly applied loss functions to indicate the trade-off between rate $R$ and distortion $D$, which can be written as:
\begin{equation}
\label{lossfunction}
\begin{aligned}
L = R + \lambda D &= R_{y} + R_{z} + R_{\text{ref}} + \lambda D \\
  &\approx R_{y} + R_{z} + \lambda D,
\end{aligned}
\end{equation}
where $R_{y}$ and $R_{z}$ represent the bit costs of the input features $\boldsymbol{y}$ and hyperprior $\boldsymbol{z}$, respectively. $R_{\text{ref}}$ indicates the bit costs of the selected indexes $i$ and $j$, which are extremely small and thus ignored in this equation. The distortion is defined as the mean square error (MSE) between the input image $\boldsymbol{x}$ and reconstructed image $\hat{\boldsymbol{x}}$, formulated as:
\begin{equation}
\label{Distortion1}
D = \frac{1}{N}\|\boldsymbol{x} - \hat{\boldsymbol{x}}\|_{2}^{2}.
\end{equation}

\begin{algorithm}[!t]
		\caption{The latent feature recycling mechanism.}
		\small
		\label{alg:freezing}
		\LinesNumbered
		\KwIn{The pre-trained REMAC network parameters  $\{\theta_{\text{E}}, \phi\}$, the input images $\{\boldsymbol{x}^{m}\}_{m=1}^{M}$, the reference images $\{\boldsymbol{x}_{\text{ref}}^{n}\}_{n=1}^{N}$, the pre-selected shallow reference images $\{\boldsymbol{x}_{\text{ref}}^{j^{m}}\}_{m=1}^{M}$.}
		\KwOut{The fine-tuned REMAC network parameters $\{\theta_{\text{E}}, \tilde{\phi}\}$.\\
			
	\textbf{Variables:} Reference features $\{\boldsymbol{y}_\text{ref}^{n}\}_{n=1}^N$, input features $\{\boldsymbol{y}^{m}\}_{m=1}^{M}$, similarity scores $\{\boldsymbol{s}_{\text{d}}^{m}\}_{m=1}^{M}$, selected indexes $\{q^{m}\}_{m=1}^{M}$. \\
    \textbf{Parameters:} Learning rate $lr$, loss $L$, max training step $K$, number of reference images $N$, number of input images $M$. 

		}
		Initialize $\tilde{\phi} \gets \phi$.    \\
\While{$n < N$}{Compute $\boldsymbol{y}_{\text{ref}}^{n}=E(\boldsymbol{x}_{\text{ref}}^{n};\theta_{\text{E}})$.
}

\While{$m < M$}{Compute $\boldsymbol{y}^{m}=E(\boldsymbol{x}^{m}; \theta_{\text{E}})$.\\
Compute $\boldsymbol{s}_{\text{d}}^{m} = \|\boldsymbol{y}^{m} - \{\boldsymbol{y}_\text{ref}^{n}\}_{n=1}^{N}\|_{1}$.\\
Compute $q^{m} = \mathop{\arg\min}\limits_{\text{index}} \boldsymbol{s}_{\text{d}}^{m}$ 
}
\For{$k=1$ to $K$}{
Compute $L$ using $\{(\boldsymbol{x}^{m}, \boldsymbol{x}_{\text{ref}}^{j^{m}}, \boldsymbol{x}_{\text{ref}}^{q^{m}})\}_{m=1}^{M}$.\\
Update $\tilde{\phi}^{k} \gets \tilde{\phi}^{k-1} + lr \times{\text{Adam}(L)}$.
}

\Return $\tilde{\phi}$.
	\end{algorithm}

\renewcommand\arraystretch{1.3} 

\begin{table*}
\footnotesize
\centering
\caption{Comparisons of the rate-distortion-complexity performance.}
\begin{threeparttable}
\resizebox{\textwidth}{!}{
\begin{tabular}{c|c|cc|cc|cc}
\toprule[1pt]
\rowcolor[gray]{0.9} 
 & &\multicolumn{6}{c}{\textbf{R-D performance}} \\
\cline{3-8}
\rowcolor[gray]{0.9} 
\multirow{-2}{*}{\textbf{Method}} &
\multirow{-2}{*}{\textbf{Enc FLOPs (G)} $\downarrow$} & \textbf{BD-PSNR (dB)} $\uparrow$ & 
\textbf{BD-rate (\%)} $\downarrow$ & 
\textbf{BD-MSSSIM (dB)} $\uparrow$ & 
\textbf{BD-rate (\%)} $\downarrow$ & 
\textbf{BD-LPIPS} $\downarrow$ & 
\textbf{BD-rate (\%)} $\downarrow$ \\
\hline

VVC \cite{vvc} & - & 0.5589 & -20.8681 & 0.6297 & -19.1038& -0.0392 & -24.3304 \\
\hline
Ball\'e \cite{hyperprior} & \textcolor{red}{151.33} & 0.0367 & -1.7809 & 0.1868 & -5.6577 & -0.0194 & -13.6237 \\
\hline
Minnen \cite{minnen} & \textcolor{blue}{362.77} & 0.5010 & -18.3098 & 0.4627 & -13.7433 & -0.0430 & -28.3645 \\
\hline
Cheng \cite{cheng} & 699.55 & \textcolor{blue}{0.6830} & \textcolor{blue}{-23.3475} &  \textcolor{blue}{0.7615} & \textcolor{blue}{-22.2249} & -0.0555 & -33.7281 \\
\hline
WACNN \cite{wacnn} & 1297.64 & 0.6560 & -21.1571 & 0.6546 & -18.3739 & \textcolor{blue}{-0.0935} & \textcolor{blue}{-47.6364} \\
\hline
Ours & 395.16 & \textcolor{red}{0.9494} & \textcolor{red}{-30.4268} & \textcolor{red}{0.8198} &  \textcolor{red}{-22.2335} & \textcolor{red}{-0.1163} & \textcolor{red}{-55.9240} \\
\bottomrule[1pt]
\end{tabular}
}
 \begin{tablenotes}
    \footnotesize
    \item[*] The FLOPs are measured under compression of 1600$\times$1152 images. Moreover, the FLOPs of similarity calculation are included in our measurement. 
    \item[**] \textcolor{red}{Red} color indicates the best result and the \textcolor{blue}{blue} color indicates the second best result.

\end{tablenotes}
      
\end{threeparttable}
\label{table:2}

\end{table*}

\begin{figure}[!t]
    \centering

    \begin{subfigure}
        \centering
        \includegraphics[width=0.5\textwidth]{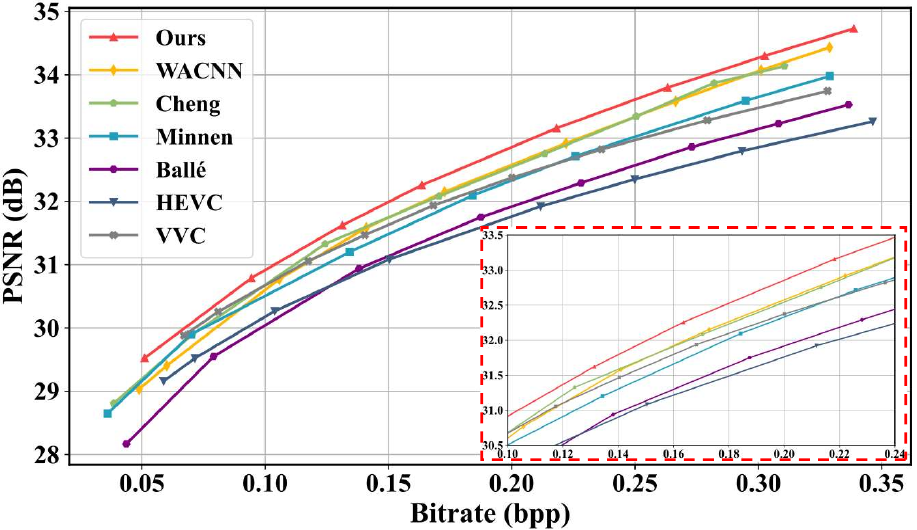}
    \end{subfigure}
    \hfill
    \begin{subfigure}
        \centering
        \includegraphics[width=0.5\textwidth]{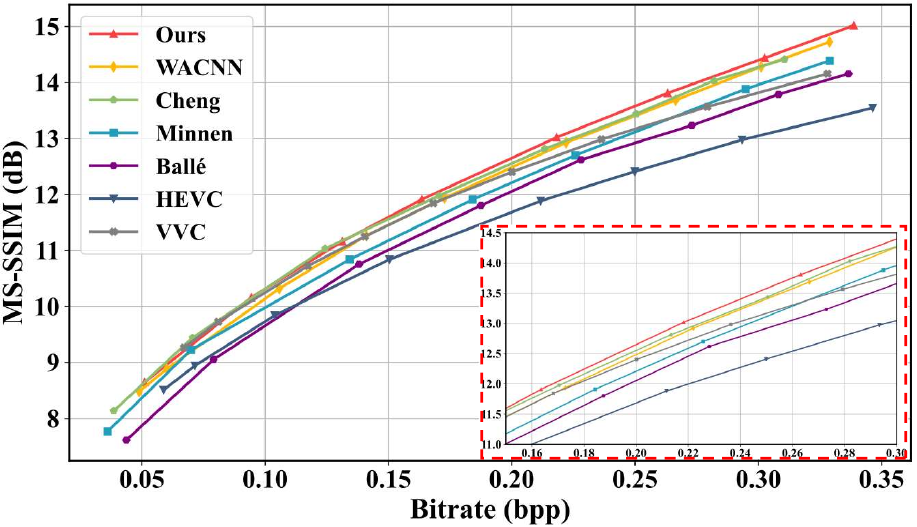}
    \end{subfigure}
    \hfill
    \begin{subfigure}
        \centering
        \includegraphics[width=0.5\textwidth]{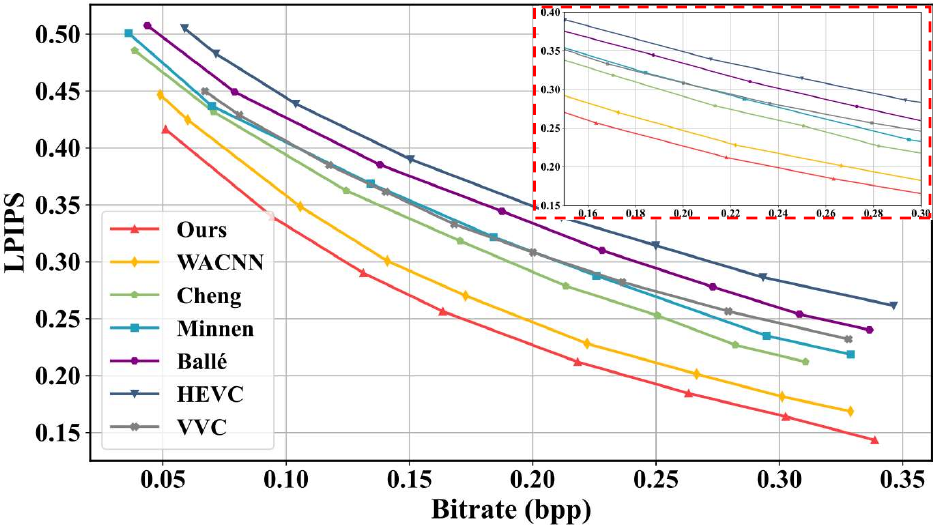}
    \end{subfigure}

    \caption{The R-D curves of our REMAC model and other compression methods in terms of PSNR, MS-SSIM, and LPIPS.}
    \label{fig:8}
\end{figure}

\section{Experiment}
\subsection{Experimental Settings}
\textit{1) Dataset:} To train the proposed REMAC framework, we adopt the MIC dataset \cite{vcip}, as it is the only dataset containing a large number of colorful and lossless Martian images. To enable reference-based image compression, we first establish a reference image dataset and construct input-reference triplets for training and validation. More specifically, the training set of $3,088$ raw images is randomly split into a $2702$-image training set and a $386$-image reference set. During the training phase, we crop the newly training and reference images into non-overlapping $512\times{512}$ patches (discarding incomplete boundary regions) and form the input-reference patch triplets based on calculated similarity scores. During the validation stage, we adopt the validation set of MIC dataset \cite{vcip} and use the uncropped reference images as reference.

\textit{2) Training:} We implement the proposed REMAC framework in CompressAI platform \cite{compressai}. To save computational resources on the encoder side, we set the channel number of latent $\boldsymbol{y}$ and hyper latent $\boldsymbol{z}$ to 192 and 128, respectively. In the proposed reference-guided entropy module, we set $M = 6$, indicating that the latent representation $\boldsymbol{y}$ is divided into 6 slices for slice-wise entropy coding. Meanwhile, we set the lambda value $\lambda$ in Eq. \ref{lossfunction} to \{0.001, 0.002, 0.003, 0.004, 0.006, 0.008, 0.010, 0.012\} for training multiple models, corresponding to various bpp (bit per pixel) points. During the training phase, we adopt the Adam \cite{Adam} optimizer and set the initial learning rate to 0.0001. Furthermore, the \textit{ReduceLROnPlateau} algorithm is employed to automatically adjust the learning rate when plateaus are detected in the validation loss.

\textit{3) Evaluation Metrics:} To comprehensively evaluate the quality of reconstructed images, we adopt two objective quality metrics and one perceptual quality metric in the following experiments. First, we utilize the peak signal-to-noise ratio (PSNR), a widely used metric in compression methods, whereby higher values denote better pixel-level fidelity. Then, we employ the MS-SSIM \cite{ms-ssim} in which higher values indicate better structural fidelity. In addition, to amplify the difference, the MS-SSIM values are represented on dB scales using $-10\times \log_{10}(1-x)$. Finally, we adopt the LPIPS \cite{lpips}, a widely used perceptual quality metric, whereby lower values correspond to better perceptual quality. To further evaluate the compression performance, we report the Bj\o ntegaard delta PSNR (BD-PSNR) \cite{bdpsnr}, BD-MSSSIM, BD-LPIPS, and BD-rate to assess the R-D performance. The BD-rate specifically measures the bit reduction achieved while maintaining image quality, which is evaluated using PSNR, MS-SSIM \cite{ms-ssim}, and LPIPS \cite{lpips}. All BD metrics reported below are computed with HEVC \cite{hevc} serving as the anchor.

\subsection{Comparisons with the State-of-the-Art Methods}
In this section, we conduct comparison experiments between traditional and LIC methods. The traditional methods consist of HEVC \cite{hevc} and VVC \cite{vvc}, which are commonly used traditional codecs. We convert the RGB images to the YCbCr color space with a 4:4:4 sampling ratio and use HEVC and VVC with their intra-coding configurations to compress the converted images. The LIC methods contain Ball\'e \cite{hyperprior}, Minnen \cite{minnen}, Cheng \cite{cheng}, and WACNN \cite{wacnn} compression models. For fair comparisons, we re-train the comparing models on the MIC \cite{vcip} dataset, which is the same dataset we used.

\textbf{Rate-Distortion-Complexity Performance.} For practical applications on Mars, we may need to balance the rate, distortion, and computational costs of the encoder. We refer to this balance as rate-distortion-complexity performance. To comprehensively evaluate the performance, we employ BD-PSNR \cite{bdpsnr}, BD-MSSSIM, and BD-LPIPS, along with their corresponding BD-rates, to measure rate-distortion performance. Meanwhile, the computational complexity of the encoder is assessed through the floating-point operations (FLOPs). We present the comparisons with the state-of-the-art compression methods in Tab. \textcolor{red}{\ref{table:2}}. Since VVC \cite{vvc} is a traditional codec, it is difficult to obtain its encoder complexity. Therefore, its encoder complexity is excluded from this table. As shown, the proposed REMAC achieves superior compression performance across PSNR, MS-SSIM \cite{ms-ssim}, and LPIPS \cite{lpips} metrics, while maintaining the third-lowest encoder complexity.
\begin{figure*}[t]
	\centering
	\includegraphics[width=7.1in]{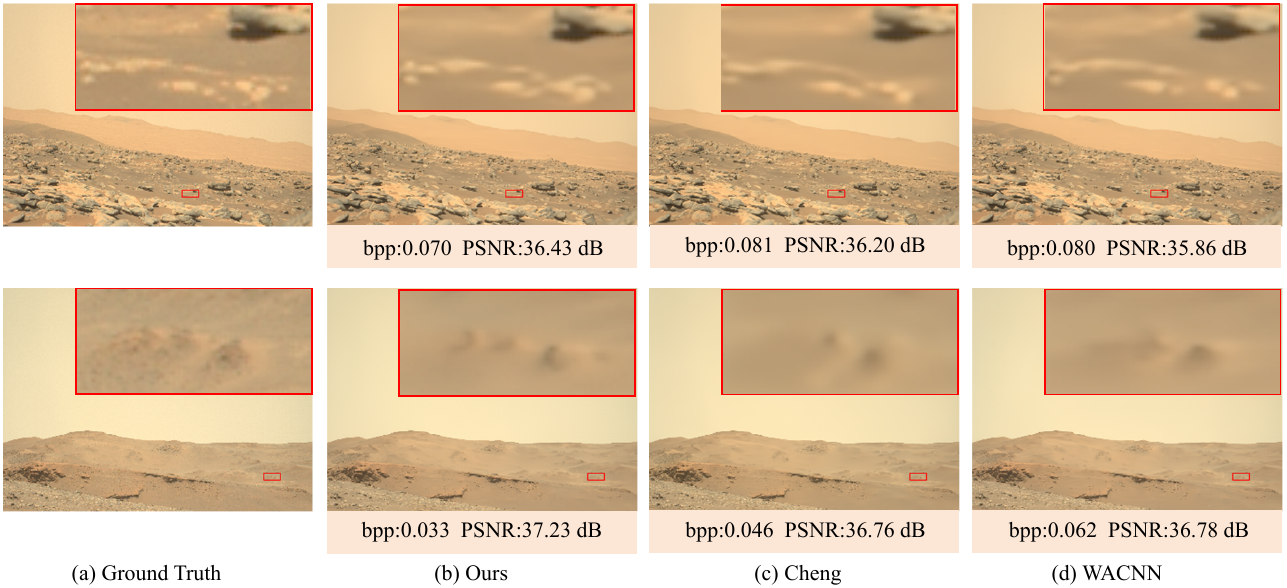}
 
	\caption{Visual comparisons of the reconstructed images.}
	\label{fig:visual}
\end{figure*}

Compared to Ball\'e \cite{hyperprior} that achieves the lowest encoding complexity, our REMAC framework achieves significant improvements of +0.9127 dB, +0.6330 dB, and -0.0969 in BD-PSNR, BD-MSSSIM, and BD-LPIPS metrics, respectively. Regarding Cheng \cite{cheng} method that achieves the second-best performance in the BD-PSNR and BD-MSSSIM metrics, the encoder complexity of our REMAC framework is only 56.49\% of that method. Compared to the WACNN \cite{wacnn} method, which achieves the second-best performance in the BD-LPIPS metric, our REMAC framework's encoder complexity is only 30.45\% of that method. Therefore, by balancing rate, distortion, and encoder complexity, our REMAC framework achieves the best rate-distortion-complexity performance.

\textbf{Rate-Distortion Performance.} 
To intuitively compare the rate-distortion performance of different methods, the R-D curves, obtained on the test set of MIC dataset \cite{vcip}, are plotted in Fig. \textcolor{red}{\ref{fig:8}}. From this figure, most LIC methods achieve superior rate-distortion performance compared to the conventional HEVC \cite{hevc} codec. This phenomenon can be attributed to the inherent \textit{inter-image} similarity among Martian images. Such similarity enables LIC methods, which are good at exploring useful and common features through CNNs, to outperform traditional codecs in rate-distortion performance. Furthermore, these R-D curves illustrate that our REMAC framework generally outperforms the state-of-the-art methods across multiple bit-rate settings, when measured by the PSNR, MS-SSIM, and LPIPS metrics.

\textbf{Visual Quality.} To facilitate intuitive comparisons of reconstruction quality, we compare reconstructed examples generated by our REMAC, and the baseline methods Cheng \cite{cheng} and WACNN \cite{wacnn} in Fig. \textcolor{red}{\ref{fig:visual}}. As shown in the visualizations, our approach achieves superior PSNR performance and lower bpps values compared to the other methods. Moreover, our REMAC preserves more texture details. From Fig. \textcolor{red}{\ref{fig:visual}}, in the red region marked on the first-row image, our REMAC preserves more details of the stone. In the second-row image, our REMAC correctly reconstructs all three mounds present in the ground truth, while both Cheng \cite{cheng} and WACNN \cite{wacnn} recover only two. The above results demonstrate the superior visual quality achieved by our framework.

\begin{figure}[t]
	\centering
	\includegraphics[width=0.48\textwidth]{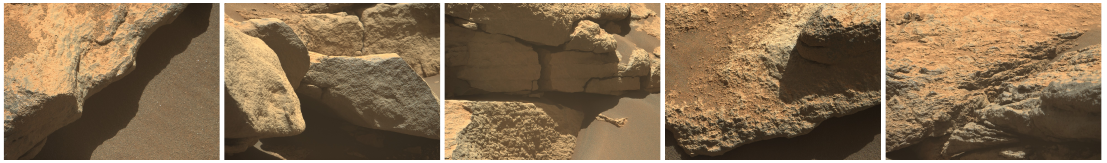}
	\caption{All five shadow-rich images from the MIC test set.}
	\label{fig:shadow}
\end{figure}

\textbf{Rate-Distortion Performance on Shadow-Rich Images.} To evaluate our method under complex lighting conditions with extensive shadows, we conduct experiments on shadow-rich images. Specifically, we define an image as shadow-rich when the shadow coverage exceeds 30\% of the total image area (\textgreater 30\%). Following this criterion, we identify five naturally occurring shadow-rich images in the test set of MIC dataset \cite{vcip}. These images all show realistic Martian shadow patterns caused by terrain structures. Fig. \textcolor{red}{\ref{fig:shadow}} shows all five shadow-rich images used in our evaluation, all of which indeed contain extensive shadows.

\begin{figure}[t]
	\centering
	\includegraphics[width=0.45\textwidth]{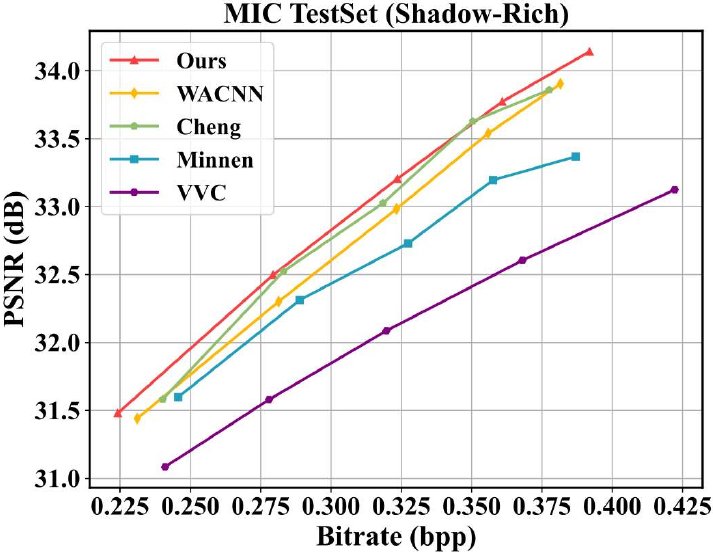}
	\caption{The R-D curves of our REMAC model and other compression methods on shadow-rich images from MIC test set.}
	\label{fig:shadow_rd}
\end{figure}

We compress these images using REMAC, along with several traditional compression and LIC methods. The R-D curves are presented in Fig. \textcolor{red}{\ref{fig:shadow_rd}}. As shown, REMAC consistently outperforms VVC \cite{vvc} across all bitrates. Moreover, REMAC achieves better compression performance than Cheng \cite{cheng}, with notably reduced encoding complexity (395.16 GFLOPs and 7.78 s compared to Cheng's 699.55 GFLOPs and 20.97 s), making it far more suitable for onboard deployment under the strict computational constraints of Mars missions. These results confirm that REMAC is both effective and practical for compressing shadow-rich images on Mars.

\textbf{Latency Cost.} To make a comprehensive comparison of inference latency between our REMAC and several strong compression baselines, we measure and report encoding and decoding latencies on both GPU and CPU in Tab. \textcolor{red}{\ref{table:timecost}}. All experiments are conducted on a fixed machine, equipped with an AMD EPYC 7642 CPU and a GeForce RTX 3090 GPU. It is important to note that the traditional codec VVC \cite{vvc} is evaluated only on CPU due to its lack of GPU acceleration. These reported latencies are averaged over 50 images, whose resolution is $1600\times1152$.

Given the Mars-to-Earth deployment scenario, where the encoder operates on Mars under strict computational constraints while the decoder runs on Earth with abundant computational resources, we prioritize encoding latency as the key efficiency metric. As shown, REMAC achieves the lowest encoding latency on both CPU and GPU, demonstrating its efficiency for on-Mars encoding. Regarding decoding, REMAC has the shortest decoding time on GPU thanks to parallel acceleration, but shows higher decoding latency on CPU, which is an expected trade-off of our asymmetrical design that shifts computational complexity from the encoder (on Mars) to the decoder (on Earth), where resources are plentiful and decoding latency does not impose a practical bottleneck.

\footnotetext[1]{The whole project detail is available at \url{https://mars.nasa.gov/mars2020/}.}

\textbf{Memory Cost.} Given the limited encoder resources of Mars rovers, we evaluate the encoder memory cost of REMAC and three compression baselines: the widely used traditional codec VVC \cite{vvc} and three representative LIC methods that achieve top performance on the MIC dataset \cite{vcip} test set, including Minnen \cite{minnen}, Cheng \cite{cheng} and WACNN \cite{wacnn}. Since current Mars rovers are equipped only with CPU-based systems, such as NASA's \textit{Perseverance} rover, all algorithms are evaluated on CPU-only hardware in a Python environment for meaningful comparisons. We evaluate them by compressing 50 randomly selected images from the test set of the MIC dataset \cite{vcip} and report the peak memory consumption during encoding. For more comprehensive comparisons, we additionally compare the compression performance (in terms of BD-PSNR) and encoding latency. Memory cost and encoder latency depend primarily on input resolution and model architecture, making averaging over 50 images provide a reliable measurement. In contrast, BD-PSNR is computed over the full test set to maintain consistency with the prior reported results in the above comparisons.

The evaluated memory costs, as well as BD-PSNR and encoder latency, are presented in Tab. \textcolor{red}{\ref{table:memorycost}}. As documented in \textit{NASA's Mars 2020 Mission}\footnotemark[1], \textit{Perseverance} rover is equipped with a BAE RAD 750 and 256 MB DRAM. To adapt to the 256 MB memory limit, we adopt a block-based compression strategy that partitions the input images to non-overlapping 320$\times$192 blocks for compression.This strategy significantly reduces memory usage. As shown in this table, block-based REMAC achieves the lowest memory cost, the second-best BD-PSNR gain, and the fastest encoder latency, demonstrating a favorable balance between resource efficiency and compression performance. Critically, under this block-based setting, REMAC consumes 207.71 MB of memory, well within the 256 MB DRAM limit of the \textit{Perseverance} rover, confirming its deployability on current Mars exploration hardware. 

\renewcommand\arraystretch{1.3} 

\begin{table}
\small
\centering
\setlength{\tabcolsep}{4pt}
\caption{Comparisons of averaged encoding and decoding latency.}
\begin{threeparttable}
\begin{tabular}{c|cc|cc}
\toprule[1pt]
\rowcolor[gray]{0.9} 
 & \multicolumn{2}{c|}{\textbf{GPU Latency (s)$\downarrow$}}& \multicolumn{2}{c}{\textbf{CPU Latency (s)$\downarrow$}} \\
\rowcolor[gray]{0.9} 
\multirow{-2}{*}{\textbf{Method}} & \textbf{Encoding} & \textbf{Decoding} & 
\textbf{Encoding}& 
\textbf{Decoding}\\
\hline

VVC \cite{vvc}& $-$ & $-$ & 177.78 & \textcolor{red}{0.32}  \\
Minnen \cite{minnen}& 15.54 &  30.63 & 23.43 & 41.39  \\
Cheng \cite{cheng}& 15.37 & 30.41 & 20.97 & 40.96  \\
WACNN \cite{wacnn}& \textcolor{blue}{0.33} & \textcolor{blue}{0.32} & \textcolor{blue}{20.58} & \textcolor{blue}{22.57}  \\
Ours & \textcolor{red}{0.17} & \textcolor{red}{0.20} & \textcolor{red}{7.78} & 50.67 
\\
\bottomrule[1pt]
\end{tabular}
 \begin{tablenotes}[normal]
    \footnotesize
    \item[*] \textcolor{red}{Red} color indicates the best result and the \textcolor{blue}{blue} color indicates the second best result.
    \item[**] Encoding latency is prioritized due to the Mars-to-Earth deployment.

\end{tablenotes}
      
\end{threeparttable}
\label{table:timecost}
\end{table}

\renewcommand\arraystretch{1.3} 

\begin{table}
\small
\setlength{\tabcolsep}{4pt}
\centering
\caption{Comparisons of encoder memory cost, BD-PSNR, and encoder latency.}
\begin{threeparttable}
\begin{tabular}{c|c|c|c}
\toprule[1pt]
\rowcolor[gray]{0.9} 
\textbf{Method}  & \textbf{Mem. (MB) $\downarrow$}&
\textbf{BD-PSNR (dB) $\uparrow$} & 
\textbf{Lat. (s) $\downarrow$} \\
\hline
VVC~\cite{vvc} & \textcolor{blue}{493.72}&
0.5589 & 
177.78 \\
Minnen~\cite{minnen}  &1106.84& 
0.5010 & 
23.43 \\
Cheng~\cite{cheng}  &961.75& 
0.6830 & 
20.97 \\
WACNN~\cite{wacnn} &1534.39& 
0.6560 & 
20.58 \\
Ours &908.52& 
\textcolor{red}{0.9494} & 
\textcolor{blue}{7.78} \\
Ours (blk) &\textcolor{red}{207.71}& 
\textcolor{blue}{0.8170} & 
\textcolor{red}{6.50} \\
\bottomrule[1pt]
\end{tabular}
\begin{tablenotes}[normal]
    \footnotesize
    \item[*] \textcolor{red}{Red} color indicates the best result and the \textcolor{blue}{blue} color indicates the second best result.
    \item[**] Ours (blk) denotes the block-based variant of our method.
 
\end{tablenotes}
\end{threeparttable}
\label{table:memorycost}
\end{table}

\begin{figure}[t]
	\centering
	\includegraphics[width=0.45\textwidth]{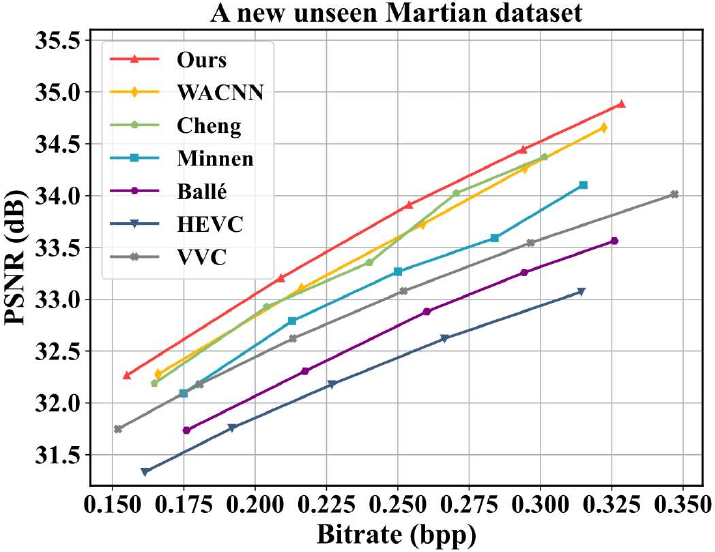}
 
	\caption{The R-D curves of our REMAC model and other compression methods on a new unseen Martian dataset.}
	\label{fig:unseen}
\end{figure}

\textbf{Generalization to Unseen Martian Data.} To evaluate generalization to unseen Martian data, we construct a new dataset based on high-resolution images from \textit{NASA's Mars 2020 Mission}\footnotemark[1] This dataset comprises 1,360 raw color images captured by the Mastcam-Z camera of the \textit{Perseverance} rover. All images are cropped to $1600\times1152$ pixels and split into training, validation, and test sets at an 8:1:1 ratio. The Fréchet Inception Distance (FID) \cite{fid} between the training sets of this new dataset and the MIC dataset \cite{vcip} (each excluding their respective held-out reference images) is 47.49, confirming that the new dataset is distributionally distinct from MIC and thus qualifies as unseen Martian data. We finetune REMAC and four representative LIC methods, including Ball\'e \cite{hyperprior}, Minnen \cite{minnen}, Cheng \cite{cheng}, and WACNN \cite{wacnn}, on this unseen dataset. Additionally, we compare against traditional compression codecs HEVC \cite{hevc} and VVC \cite{vvc}. The R-D curves on the test set are shown in Fig. \textcolor{red}{\ref{fig:unseen}}. As shown, REMAC consistently achieves superior compression performance across all bpp points, demonstrating strong generalization to unseen Martian data.

\subsection{Ablation Study}
In this section, we conduct ablation studies to validate the effectiveness of the proposed ref-hyperprior and ref-decoder. Moreover, we also conduct the ablation experiments to investigate the effectiveness of the proposed latent feature recycling mechanism.

\begin{figure}[t]
	\centering
	\includegraphics[width=0.45\textwidth]{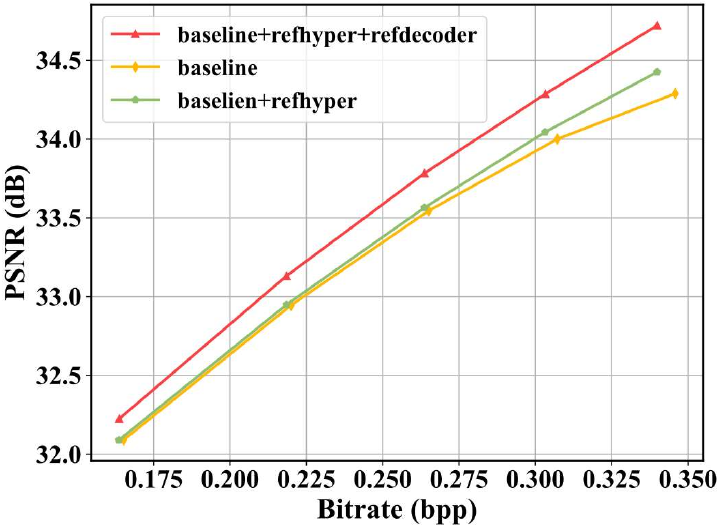}
 
	\caption{The R-D curves of the baseline, ``baseline + ref-hyperprior'',  and ``baseline + ref-hyperprior + ref-decoder'' in terms of PSNR.}
	\label{fig:ablation_remac}
\end{figure}

\textbf{Effectiveness of Ref-Hyperprior.} 
The ref-hyperprior is designed to leverage the reference features $\boldsymbol{y}_{\text{ref}}^{i}$ to obtain useful information for improving entropy estimation of input features $\boldsymbol{y}$. To verify its effectiveness, we first form a baseline model by integrating our proposed input encoder with the channel-wise entropy autoregressive model \cite{channel-wise}. Subsequently, we incorporate our proposed ref-hyperprior into the baseline model. Given that the ref-hyperprior requires the reference features as its input, we further add the proposed ref-encoder into the baseline model, obtaining the ``baseline + ref-hyperprior'' model. We train them and the corresponding R-D curves are presented in Fig. \textcolor{red}{\ref{fig:ablation_remac}}. From this figure, the ``baseline + ref-hyperprior'' model achieves obvious PSNR gains at the higher bpp points. This validates the effectiveness of our ref-hyperprior. Moreover, the baseline model obtains +0.9558 dB in BD-PSNR, while the ``baseline + ref-hyperprior'' model attains +1.0005 dB. These results further indicate the effectiveness of ref-hyperprior.

\textbf{Effectiveness of Ref-Decoder.} 
The ref-decoder is proposed to exploit the reference image $\boldsymbol{x}_{\text{ref}}^{j}$ for providing valuable information during output image reconstruction $\hat{\boldsymbol{x}}$. To validate its effectiveness, we integrate the proposed ref-decoder into the baseline model with ref-hyperprior (i.e., ``baseline + ref-hyperprior''), forming the ``baseline + ref-hyperprior + ref-decoder'' model. We train this model and present its R-D curves in Fig. \textcolor{red}{\ref{fig:ablation_remac}}. As illustrated, across all bpp points, the ``baseline + ref-hyperprior + ref-decoder'' model achieves superior PSNR compared to the ``baseline + ref-hyperprior'' model. This improvement quantifies enhanced pixel-wise fidelity in reconstructed images, thereby verifying the effectiveness of the proposed ref-decoder. Moreover, we calculate its BD-PSNR relative to HEVC. The ``baseline + ref-hyperprior + ref-decoder'' model achieves a BD-PSNR gain of +1.2016 dB. Moreover, the ``baseline + ref-hyperprior + ref-decoder'' model demonstrates a PSNR improvement of +0.2011 dB over the ``baseline + ref-hyperprior'' model. These results further indicate the effectiveness of the ref-decoder.

\begin{figure}[t]
	\centering
	\includegraphics[width=0.45\textwidth]{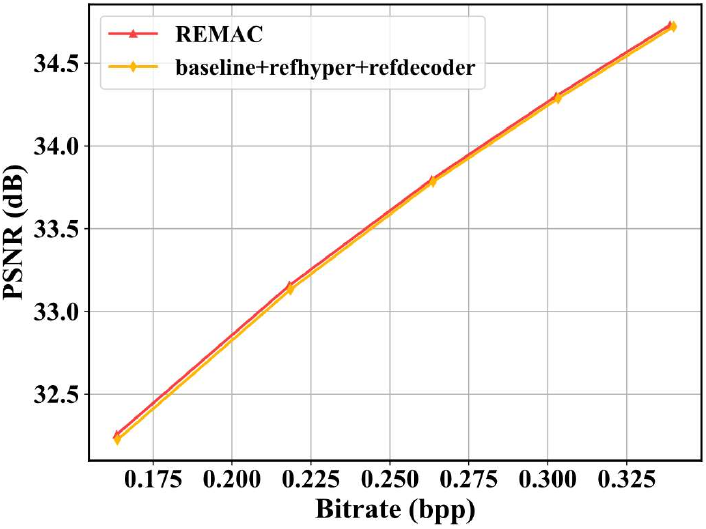}
 
	\caption{The R-D curves of the ``baseline + ref-hyperprior + ref-decoder'' and REMAC in terms of PSNR. These results are attained with FLOPs decrease.}
	\label{fig:remac_final_1009}
\end{figure}

\textbf{Effectiveness of Latent Feature Recycling Mechanism.} 
The proposed latent feature recycling mechanism is designed to significantly reduce encoder complexity without substantially compromising compression performance. To validate its effectiveness, we apply this mechanism to the ``baseline + ref-hyperprior + ref-decoder'' model, resulting in the final REMAC model. The R-D curves of both models are illustrated in Fig. \textcolor{red}{\ref{fig:remac_final_1009}}. From this figure, the REMAC model achieves superior PSNR values across all bit-rate points compared to the ``baseline + ref-hyperprior + ref-decoder'' model. Furthermore, the encoding FLOPs of the ``baseline + ref-hyperprior + ref-decoder'' model comprise the computations of the pre-encoder (shared backbone) on the input image, the input-encoder, and the input-involved components of the reference-guided entropy module. In contrast, all computations that exclusively process the reference images are excluded, as the reference features are assumed to be pre-computed and stored on the encoder side. Specifically, the excluded computations include the pre-encoder applied to the reference image, the ref-encoder, and the generation of reference-specific parameters $(\mu^{''}, \sigma^{''})$ in the reference-guided entropy module. Regarding the REMAC model's  encoder complexity, our proposed latent feature recycling mechanism reuses the deep latent features from the input-encoder for deep reference selection, thereby eliminating the computations of the pre-encoder to generate deep input features and further reducing encoder complexity. When compressing 1600$\times$1152 images, the encoding FLOPs of the ``baseline + ref-hyperprior + ref-decoder'' model amount to 406.85G, while the REMAC model requires only 395.16G. This represents a 2.87\% reduction in encoding FLOPs, making it more suitable for Martian image compression with resource-constrained encoders. These results confirm the effectiveness of the proposed latent feature recycling mechanism.

\begin{figure}[t]
	\centering
	\includegraphics[width=0.45\textwidth]{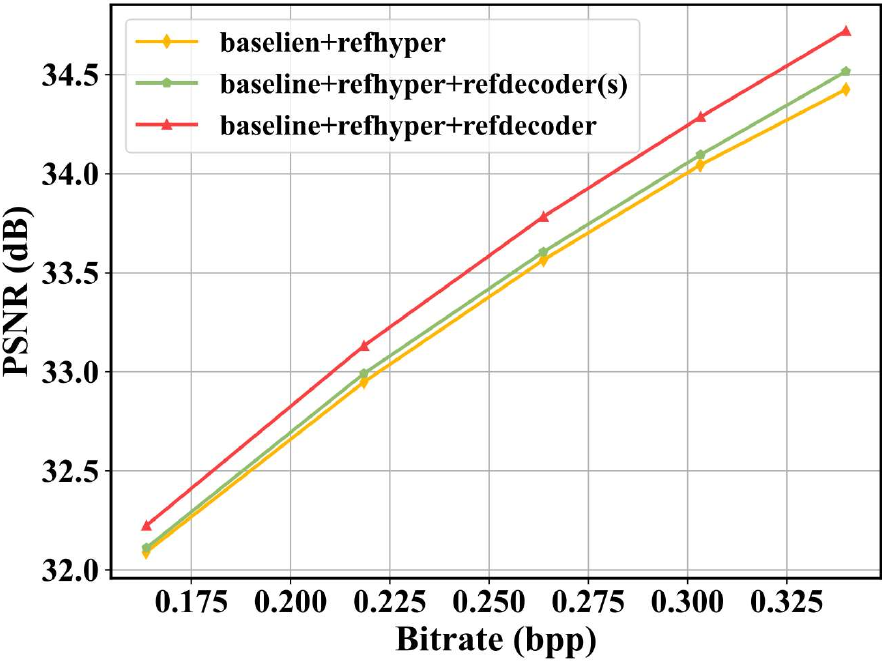}
 
	\caption{The R-D curves of the symmetrical ``baseline + ref-hyperprior'', the asymmetrical ``baseline + ref-hyperprior + ref-decoder'', and its simplified variant ``baseline + ref-hyperprior + ref-decoder(s)'' models in terms of PSNR.}
	\label{fig:assy}
\end{figure}

\textbf{Effectiveness of Asymmetrical Design.} The asymmetrical design aims to exploit the decoder's computational resources to compensate for the performance limitations of a lightweight encoder. To assess its effectiveness, we first extend the baseline model with the proposed ref-hyperprior (denoted as ``baseline + ref-hyperprior''), resulting in a symmetrical architecture. Then, we further incorporate the proposed ref-decoder to form the full asymmetrical model, which is denoted as ``baseline + ref-hyperprior + ref-decoder''. To further evaluate the impact of decoder complexity, we also introduce a simplified and asymmetrical variant, ``baseline + ref-hyperprior + ref-decoder(s)'', where the depth of the decoder is reduced by removing several convolutional layers and residual blocks, while preserving its multi-scale structure. The R-D curves of these models are shown in Fig. \textcolor{red}{\ref{fig:assy}}. As illustrated, both the full and simplified asymmetrical models consistently outperform the symmetrical baseline across all bpp points in terms of PSNR. For quantitative comparison, we compute BD-PSNR using HEVC as the anchor. The symmetrical baseline achieves a BD-PSNR gain of +1.0005 dB, while the full and simplified asymmetrical models achieve gains of +1.2016 dB and +1.0459 dB, respectively. Both the asymmetrical models achieve greater BD-PSNR gains than the symmetrical baseline, demonstrating the effectiveness of the asymmetrical design. Furthermore, the performance gap between the simplified and full asymmetrical models highlights the benefit of a more complex decoder, indicating that increased decoder capacity positively contributes to reconstruction quality.

\begin{figure*}[htbp]
    \centering
    \begin{subfigure}
        \centering
        \includegraphics[width=0.24\textwidth]{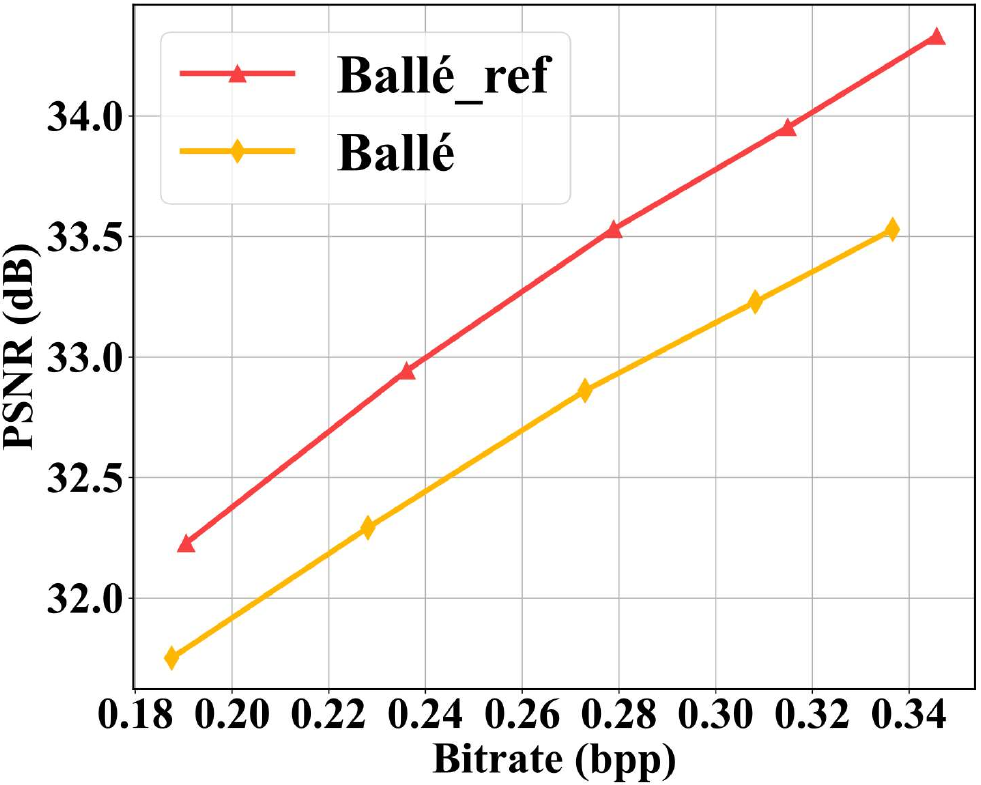}
        \label{fig:psnr_balle}
    \end{subfigure}
    \begin{subfigure}
        \centering
        \includegraphics[width=0.24\textwidth]{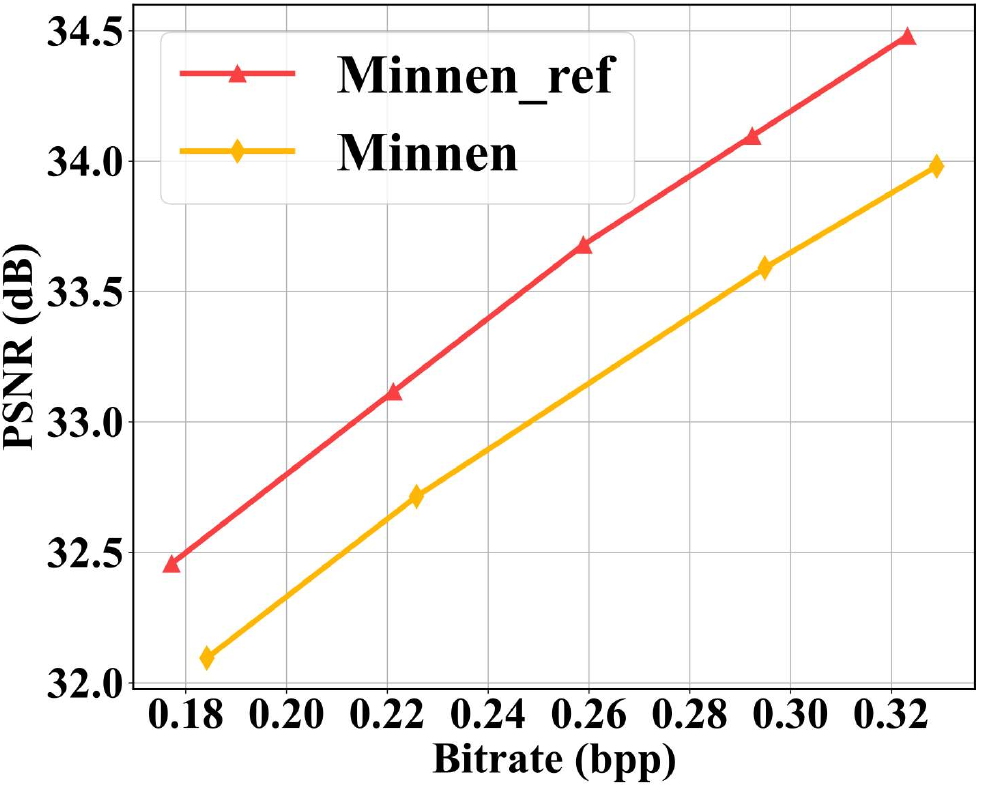}
        \label{fig:psnr_mbt}
    \end{subfigure}
    \begin{subfigure}
        \centering
        \includegraphics[width=0.24\textwidth]{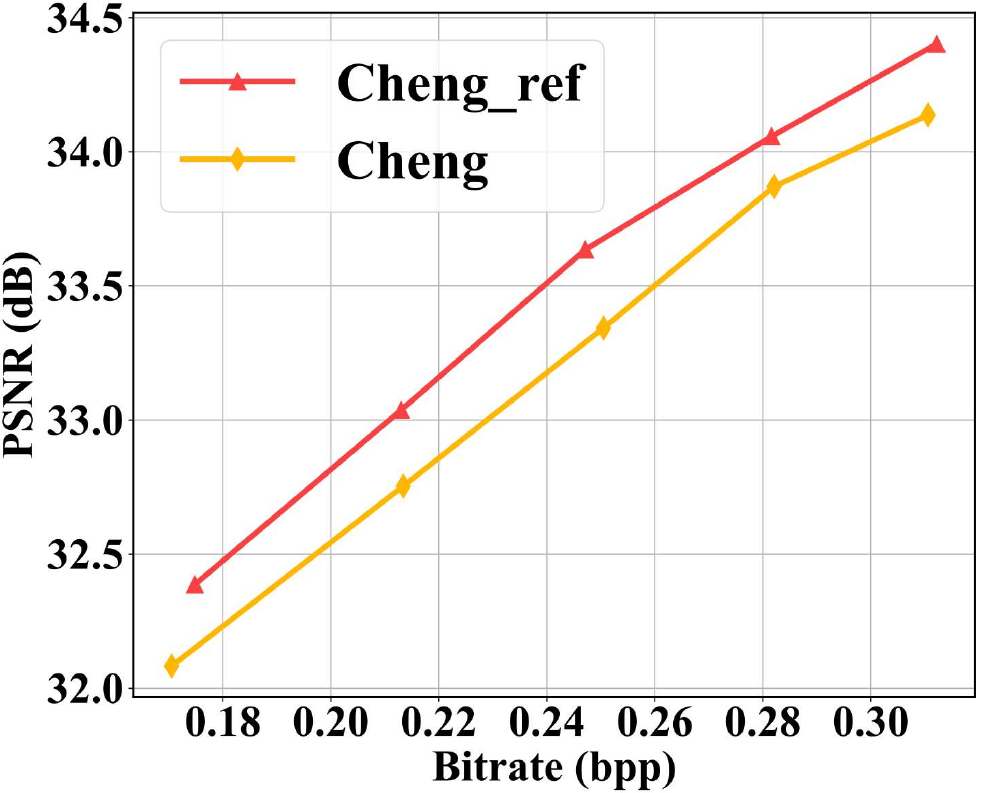}
        \label{fig:psnr_cheng}
    \end{subfigure}
    \begin{subfigure}
        \centering
\includegraphics[width=0.24\textwidth]{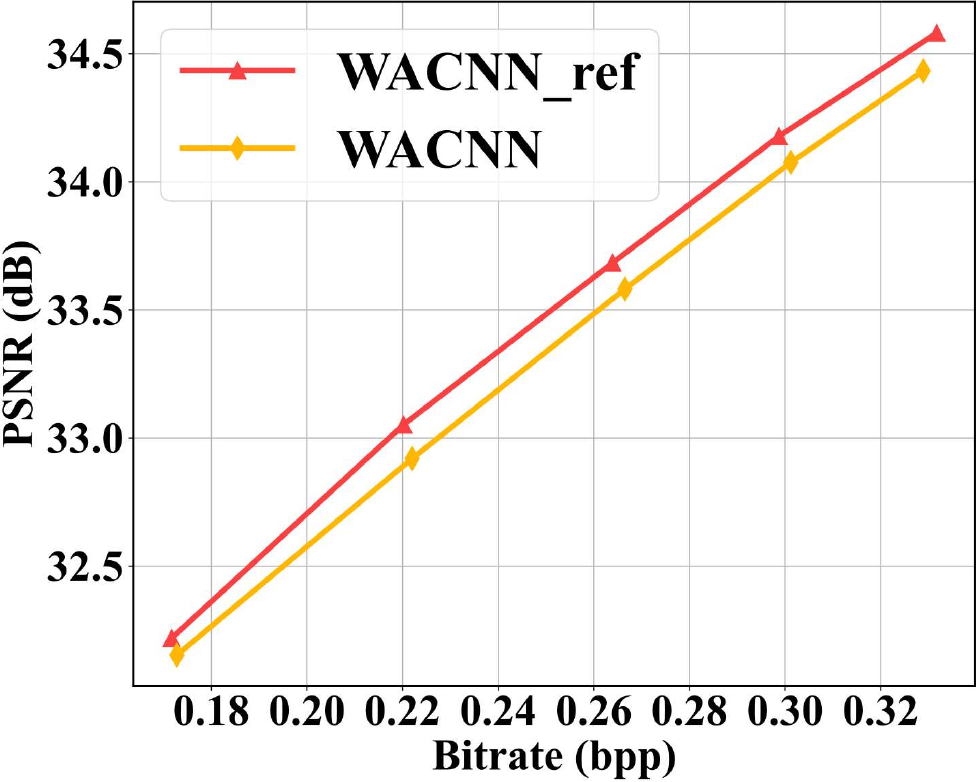}
        \label{fig:psnr_wacnn}
    \end{subfigure}
    
    \caption{The R-D curves of the existing image compression methods and their corresponding ref-enhanced variants in terms of PSNR.}
    \label{fig:ref_var_psnr}
\end{figure*}

\textbf{Effectiveness of Reference-based Mechanism.} The reference-based mechanism for Martian image compression is motivated by the high \textit{inter-image} similarity inherent in Martian images. To assess its general effectiveness, we integrate this mechanism into four representative LIC methods, including Ball\'e \cite{hyperprior}, Minnen \cite{minnen}, Cheng \cite{cheng}, and WACNN \cite{wacnn} compression models. For each baseline model, we introduce components analogous to our proposed reference-guided entropy module and ref-decoder, with necessary adaptations made for each network architecture. Each adaptation was carefully tailored to preserve the original architecture's integrity, such as preserving $5\times5$ convolutions in WACNN. The R-D curves of the original models and their reference-enhanced variants are shown in Fig. \textcolor{red}{\ref{fig:ref_var_psnr}}. Across all bpp points, every reference-enhanced variant achieves consistent PSNR gains, confirming that explicitly modeling \textit{inter-image} similarity is a robust and architecture-agnostic strategy for enhancing compression efficiency on Martian images.

\textbf{Robustness of Reference Selection.} To evaluate the robustness of our feature-level similarity-based reference selection, we conduct comprehensive experiments to assess its robustness to scene variations, sensor noise, and compression artifacts. 

\begin{figure*}[htbp]
    \centering
    \begin{subfigure}
        \centering
        \includegraphics[width=0.24\textwidth]{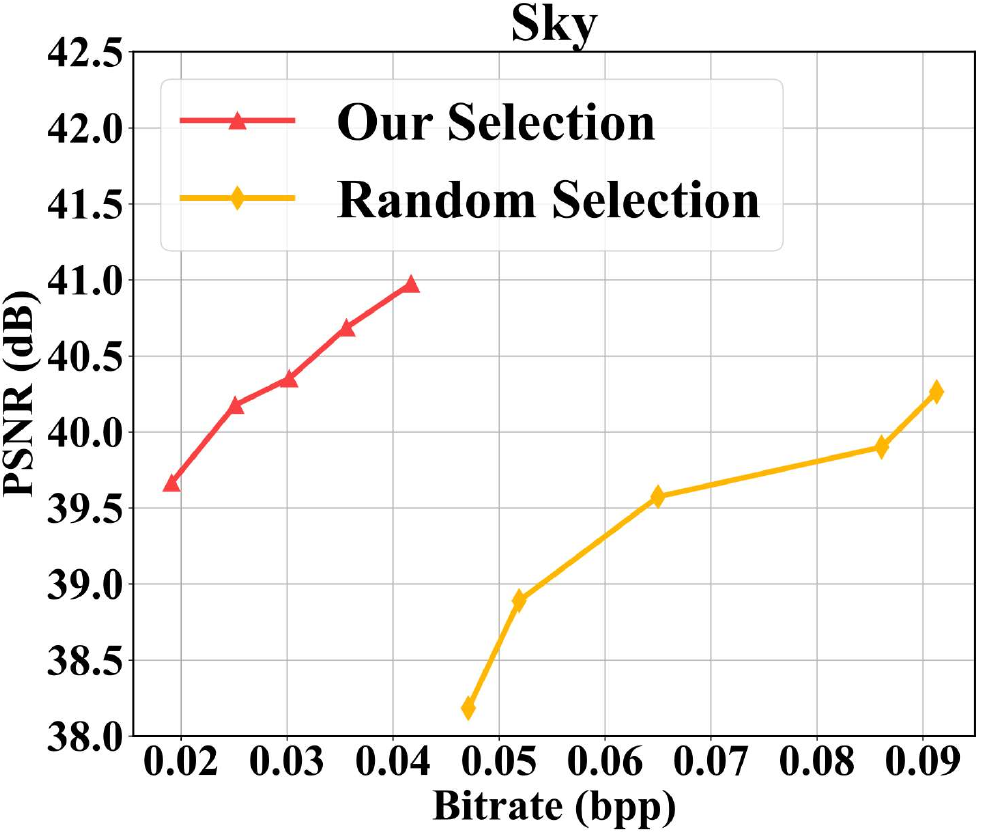}
        \label{fig:psnr_balle}
    \end{subfigure}
    \begin{subfigure}
        \centering
        \includegraphics[width=0.24\textwidth]{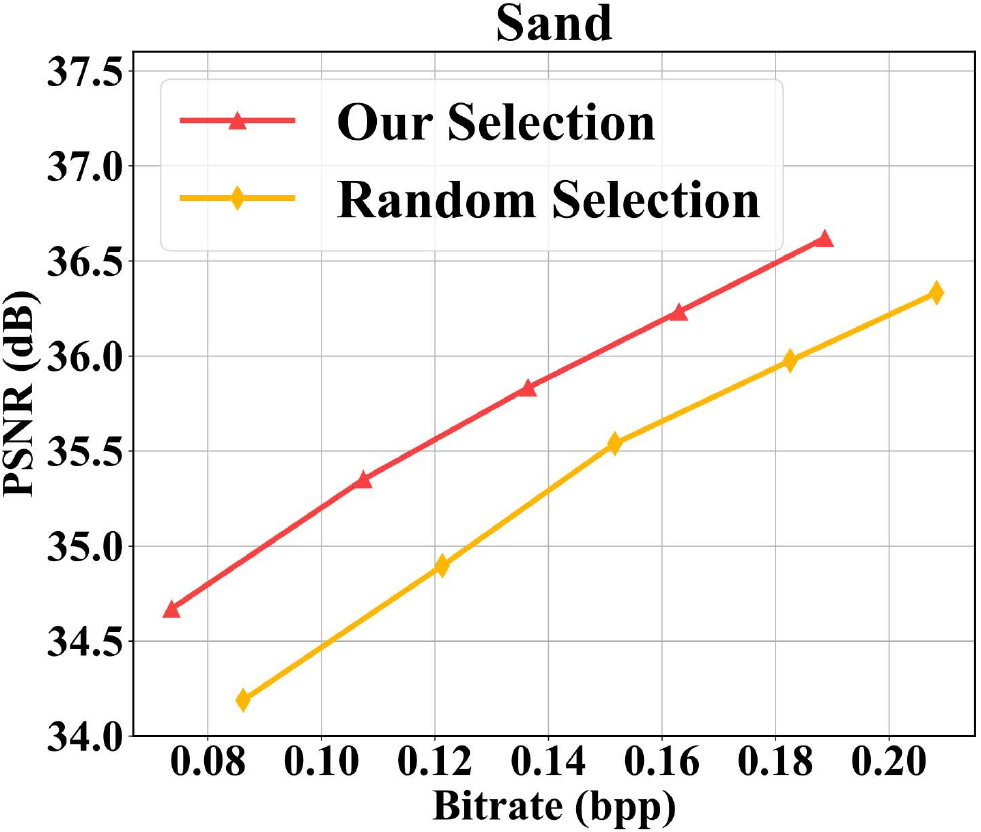}
        \label{fig:psnr_mbt}
    \end{subfigure}
    \begin{subfigure}
        \centering
        \includegraphics[width=0.24\textwidth]{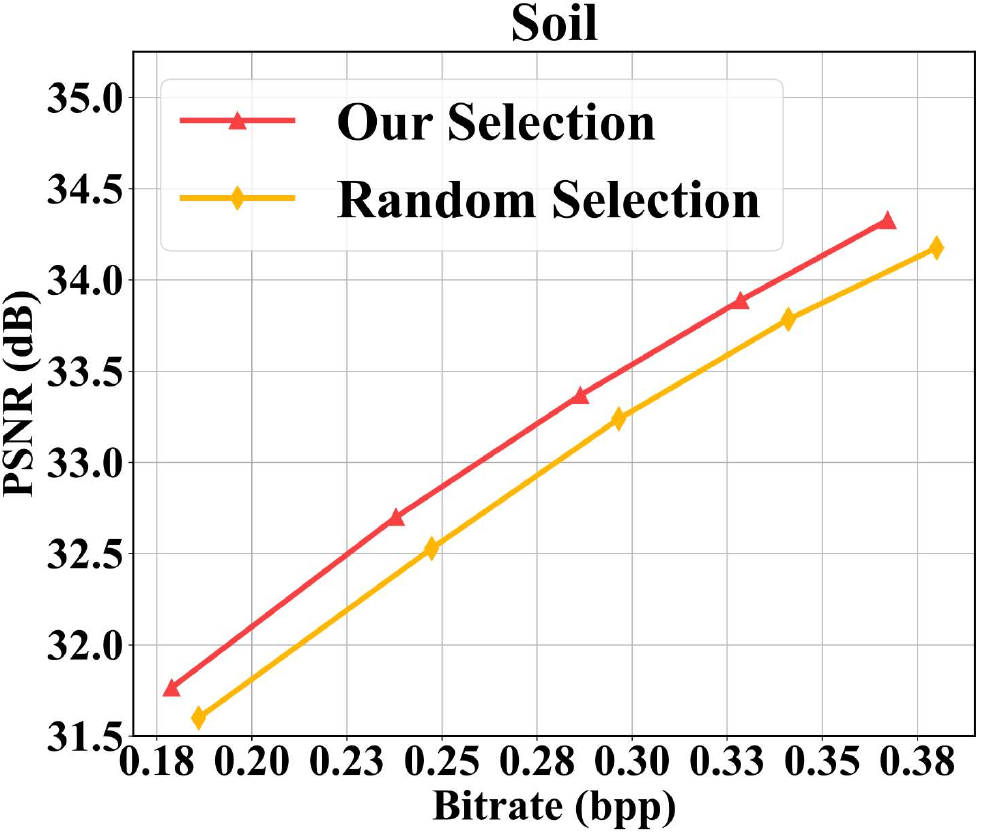}
        \label{fig:psnr_cheng}
    \end{subfigure}
    \begin{subfigure}
        \centering
\includegraphics[width=0.24\textwidth]{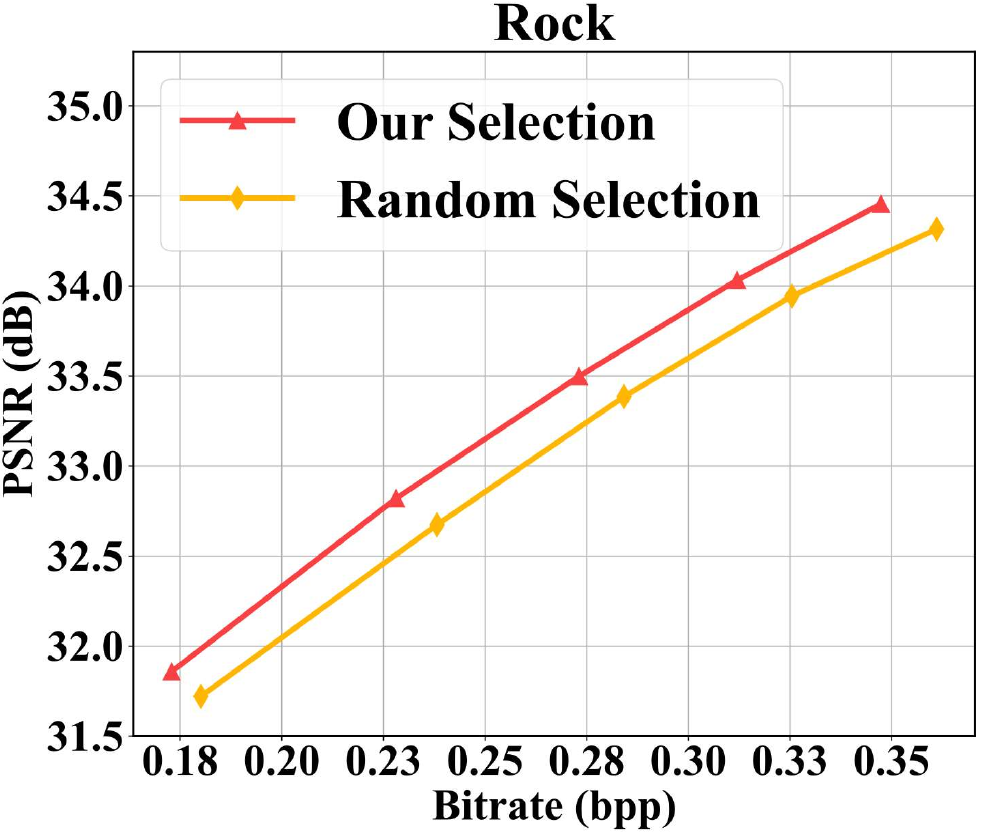}
        \label{fig:psnr_wacnn}
    \end{subfigure}
  
    \caption{The R-D curves of our feature-level similarity-based selection and random selection on four categories, including sky, sand, soil, and rock, in terms of PSNR.}
    \label{fig:crossScene}
\end{figure*}

1) Robustness to scene variations: To evaluate cross-scene robustness, we partition the MIC dataset \cite{vcip} test set into four scene categories: sky, sand, soil, and rock. For each category, we compare the compression performance of two reference selection strategies: (i) our feature-level similarity-based selection and (ii) random selection. To ensure statistical reliability, we repeat the random reference selection 10 times independently, each time constructing new input-reference triplets, and report average results. As shown in Fig. \textcolor{red}{\ref{fig:crossScene}}, our feature-level similarity-based selection consistently achieves better R-D performance than the random selection across all scene types in terms of PSNR, demonstrating its robustness to scene variations.

\begin{table*}
\small
\centering
\caption{Accuracy of shallow and deep reference selection with different Gaussian noise levels.}
\begin{threeparttable}
\begin{tabular}{c|c|ccccc}
\toprule[1pt]
\rowcolor[gray]{0.9} 
 & \textbf{Shallow Selection (\%)}& \multicolumn{5}{c}{\textbf{Deep Selection (\%)}} \\
\rowcolor[gray]{0.9} 
\multirow{-2}{*}{\textbf{Noise Level}} &  $\lambda \in \{0.012, 0.010, 0.008, 006\}$ & $\lambda=0.012$ & $\lambda=0.010$&$\lambda=0.008$&$\lambda=0.006$&Average\\
\hline
$\sigma = 5$ & 96.89 & 97.15 & 95.34 & 93.01&93.78&94.82  \\
$\sigma = 10$& 95.60 & 94.04 & 91.45& 91.45&91.97&92.23  \\
$\sigma = 15$& 95.85 & 90.93 & 88.60 & 89.12&90.16&89.70  \\
\bottomrule[1pt]
\end{tabular}
 \begin{tablenotes}[normal]
    \footnotesize
    \item[*] In shallow selection, the pretrained features are adopted to similarity calculation, thus not affected by $\lambda$ of REMAC.
    \item[**] In deep selection, the features of R-Encoder are reused to similarity calculation, thus affected by $\lambda$ of REMAC.

\end{tablenotes}
      
\end{threeparttable}
\label{table:GaussianNoise}
\end{table*}

\begin{table*}
\small
\centering
\caption{Accuracy of shallow and deep reference selection with different JPEG quality factors.}
\begin{threeparttable}
\begin{tabular}{c|c|ccccc}
\toprule[1pt]
\rowcolor[gray]{0.9} 
 & \textbf{Shallow Selection (\%)}& \multicolumn{5}{c}{\textbf{Deep Selection (\%)}} \\
\rowcolor[gray]{0.9} 
\multirow{-2}{*}{\textbf{Noise Level}} &  $\lambda \in \{0.012, 0.010, 0.008, 006\}$ & $\lambda=0.012$ & $\lambda=0.010$&$\lambda=0.008$&$\lambda=0.006$&Average\\
\hline
QF = 50 & 97.93 & 97.67 & 98.19 & 97.15&97.93&97.74  \\
QF = 75& 98.70 & 98.70 & 97.41& 98.45&98.70&98.32  \\
QF = 90& 98.96 & 99.48 & 99.22 & 99.22&99.48&99.35  \\
\bottomrule[1pt]
\end{tabular}
 \begin{tablenotes}[normal]
    \footnotesize
    \item[*] In shallow selection, the pretrained features are adopted to similarity calculation, thus not affected by $\lambda$ of REMAC.
    \item[**] In deep selection, the features of R-Encoder are reused to similarity calculation, thus affected by $\lambda$ of REMAC.

\end{tablenotes}
      
\end{threeparttable}
\label{table:JPEGNoise}
\end{table*}

2) Robustness to sensor noise: To validate robustness to sensor noise, we add Gaussian noise with $\sigma \in \{5, 10, 15\}$ to the MIC dataset \cite{vcip} test set, simulating varying levels of sensor degradation. In our dual-reference selection strategy, one reference is selected based on shallow texture features (shallow selection), and another based on deep semantic features (deep selection). We report the accuracy of each selection method relative to the noise-free baseline to assess robustness. Results are reported in Tab. \textcolor{red}{\ref{table:GaussianNoise}}. As shown, across all noise levels, the accuracy of shallow selection remains above 95.85\%, while the average accuracy of deep selection stays above 89.70\%, demonstrating strong resilience to sensor noise. This indicates that both shallow and deep selections are robust to sensor noise. 

3) Robustness to compression artifacts: To validate the robustness to compression artifacts, we use the widely adopted codec JPEG \cite{jpeg} as a proxy to simulate common compression artifacts. Specifically, we apply JPEG compression with quality factors $QF \in \{50, 75, 90\}$ to the MIC dataset \cite{vcip} test set. We evaluate the accuracy of both shallow and deep reference selection relative to the artifact-free baseline. Results are presented in Tab. \textcolor{red}{\ref{table:JPEGNoise}}. As shown, even at the lowest quality factor ($QF = 50$), the accuracy of shallow selection remains above 97.93\%, while the average accuracy of deep selection exceeds 97.74\%. This high resilience indicates that both shallow texture and deep semantic features retain sufficient discriminative power under severe blocky distortion. The minimal degradation confirms that our dual-reference selection strategy is robust to compression artifacts.

\textbf{Generalization to Input-Reference Mismatch.} To evaluate the generalization capability of REMAC to the input-reference mismatch, we conduct additional experiments using the following three reference selection strategies that simulate increasing levels of difference between input and reference images: (1) our feature-level similarity-based strategy with the held-out reference set derived from the MIC training set \cite{vcip} (mild difference), (2) random selection from the same reference set (moderate difference), and (3) random selection from an out-of-distribution reference set consisting of unseen images from \textit{NASA's Mars 2020 Mission}\footnotemark[1] (significant difference). This represents a highly challenging and realistic mismatch scenario. The FID \cite{fid} between this newly constructed Mars reference set and the actual training set used to train REMAC (i.e., the MIC training set excluding the held-out reference images) is 62.54, confirming a substantial distributional shift. In contrast, the FID between the actual training set and the held-out reference set is 33.98, substantially lower than 62.54, indicating a small but non-negligible distributional gap that underlies both the mild and moderate mismatch scenarios. These settings allow us to systematically analyze REMAC's generalization capability across mild, moderate, and significant input-reference mismatch scenarios. 

\begin{figure}[t]
	\centering
	\includegraphics[width=3.2in]{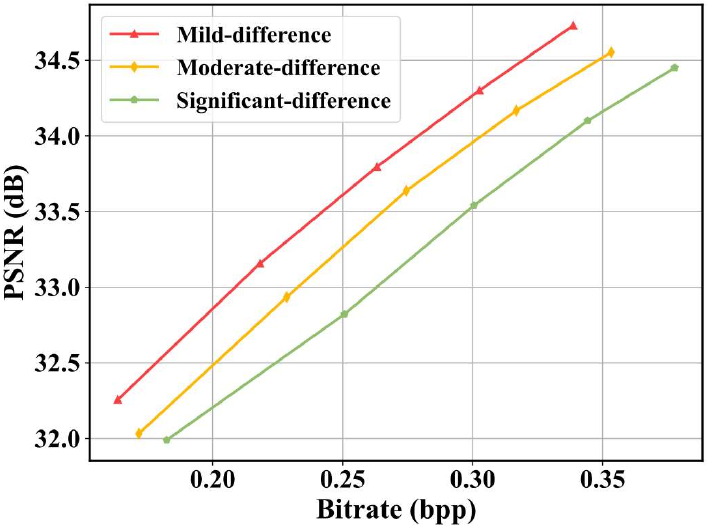}
	\caption{The R-D curves of different levels of input-reference mismatch.}
	\label{fig:mild}
\end{figure}

Specifically, to construct an out-of-distribution reference set, we collect 386 unseen Martian images from \textit{NASA's Mars 2020 Mission}\footnotemark[1], matching the size of our original reference set. To ensure statistical reliability, we repeat the random reference selection process $10$ times independently, each time constructing new input-reference images, and report the average compression results in Fig. \textcolor{red}{\ref{fig:mild}}. From this figure, the mild-difference scenario (using our feature-level similarity-based selection) consistently achieves the best R-D performance across all bpp points in terms of PSNR. Meanwhile, with the input-reference mismatch ranging from mild to significant, the R-D performance gradually decreases. The PSNR drops are marginal, indicating that REMAC consistently reconstructs images that closely resemble the originals and has the generalization capability under input-reference mismatch scenarios.

\section{Limitations and Future Work}
While REMAC achieves state-of-the-art R-D performance compared with the existing compression methods, its current direct implementation incurs large memory and computational demands, primarily due to the use of full-precision (32-bit floating-point) weights and the intensive inference process inherent in deep learning methods. This limitation highlights the need for model efficiency improvements. In future work, we plan to explore lightweight strategies, such as weight quantization and model pruning, to reduce its memory and computational costs while preserving compression performance.

\section{Conclusion}
In this paper, we have proposed the REMAC approach for effective and efficient Martian image compression. Motivated by an empirical analysis of 784 Martian images, revealing strong \textit{intra-} and \textit{inter-image} similarities in terms of color, texture, and semantics, we designed REMAC to exploit these similarities for improved compression performance and reduced encoder complexity. By leveraging reference images available at both the encoder and decoder, REMAC shifts computational complexity from the resource-limited encoder to the resource-rich decoder. To exploit the \textit{inter-image} similarities, we proposed a reference-guided entropy module and a ref-decoder that enhance both entropy estimation and reconstruction quality using valuable reference information. To leverage the \textit{intra-image} similarities, the ref-decoder adopts a deep, multi-scale architecture with a large receptive field size to model long-range spatial dependencies within individual images. Furthermore, a latent feature recycling mechanism was developed to reuse pre-extracted features, reducing specific computation during reference image selection. Finally, extensive experiments were conducted to demonstrate the superiority of our REMAC approach in the rate-distortion-complexity performance. 

\bibliography{TGRS}

\begin{thebibliography}{10}
\providecommand{\url}[1]{#1}
\csname url@samestyle\endcsname
\providecommand{\newblock}{\relax}
\providecommand{\bibinfo}[2]{#2}
\providecommand{\BIBentrySTDinterwordspacing}{\spaceskip=0pt\relax}
\providecommand{\BIBentryALTinterwordstretchfactor}{4}
\providecommand{\BIBentryALTinterwordspacing}{\spaceskip=\fontdimen2\font plus
\BIBentryALTinterwordstretchfactor\fontdimen3\font minus \fontdimen4\font\relax}
\providecommand{\BIBforeignlanguage}[2]{{%
\expandafter\ifx\csname l@#1\endcsname\relax
\typeout{** WARNING: IEEEtran.bst: No hyphenation pattern has been}%
\typeout{** loaded for the language `#1'. Using the pattern for}%
\typeout{** the default language instead.}%
\else
\language=\csname l@#1\endcsname
\fi
#2}}
\providecommand{\BIBdecl}{\relax}
\BIBdecl

\bibitem{MarsExp1}
S.~Liu, Y.~Lin, K.~Du, J.~Zhang, X.~Tong, H.~Xie, B.~Wan, and Z.~Zhang, ``A novel in situ dust cover index for analyzing the multispectral camera image acquired by china’s zhurong mars rover,'' \emph{IEEE Transactions on Geoscience and Remote Sensing}, vol.~63, pp. 1--13, 2025.

\bibitem{MarsExp2}
T.~Panambur and M.~Parente, ``Enhancing martian terrain recognition with deep constrained clustering,'' \emph{IEEE Transactions on Geoscience and Remote Sensing}, pp. 1--1, 2025.

\bibitem{MarsExp3}
X.~Tan, B.~Xi, H.~Xu, J.~Li, Y.~Li, C.~Xue, and J.~Chanussot, ``A lightweight framework with knowledge distillation for zero-shot mars scene classification,'' \emph{IEEE Transactions on Geoscience and Remote Sensing}, vol.~62, pp. 1--16, 2024.

\bibitem{kim2020efficient}
J.-H. Kim, J.-H. Choi, J.~Chang, and J.-S. Lee, ``Efficient deep learning-based lossy image compression via asymmetric autoencoder and pruning,'' in \emph{Icassp 2020-2020 ieee international conference on acoustics, speech and signal processing (icassp)}.\hskip 1em plus 0.5em minus 0.4em\relax IEEE, 2020, pp. 2063--2067.

\bibitem{hevc}
G.~J. Sullivan, J.-R. Ohm, W.-J. Han, and T.~Wiegand, ``Overview of the high efficiency video coding (hevc) standard,'' \emph{IEEE Transactions on Circuits and Systems for Video Technology}, vol.~22, no.~12, pp. 1649--1668, 2012.

\bibitem{vvc}
B.~Bross, Y.-K. Wang, Y.~Ye, S.~Liu, J.~Chen, G.~J. Sullivan, and J.-R. Ohm, ``Overview of the versatile video coding (vvc) standard and its applications,'' \emph{IEEE Transactions on Circuits and Systems for Video Technology}, vol.~31, no.~10, pp. 3736--3764, 2021.

\bibitem{vcip}
Q.~Ding, M.~Xu, S.~Li, X.~Deng, Q.~Shen, and X.~Zou, ``A learning-based approach for martian image compression,'' in \emph{2022 IEEE International Conference on Visual Communications and Image Processing (VCIP)}, 2022, pp. 1--5.

\bibitem{cheng}
Z.~Cheng, H.~Sun, M.~Takeuchi, and J.~Katto, ``Learned image compression with discretized gaussian mixture likelihoods and attention modules,'' in \emph{2020 IEEE/CVF Conference on Computer Vision and Pattern Recognition (CVPR)}, 2020, pp. 7936--7945.

\bibitem{wacnn}
R.~Zou, C.~Song, and Z.~Zhang, ``The devil is in the details: Window-based attention for image compression,'' in \emph{2022 IEEE/CVF Conference on Computer Vision and Pattern Recognition (CVPR)}, 2022, pp. 17\,471--17\,480.

\bibitem{hyperprior}
J.~Ball{\'e}, D.~Minnen, S.~Singh, S.~J. Hwang, and N.~Johnston, ``Variational image compression with a scale hyperprior,'' \emph{arXiv preprint arXiv:1802.01436}, 2018.

\bibitem{balle1}
J.~Ball{\'e}, V.~Laparra, and E.~P. Simoncelli, ``End-to-end optimized image compression,'' \emph{arXiv preprint arXiv:1611.01704}, 2016.

\bibitem{ViT}
A.~Dosovitskiy, L.~Beyer, A.~Kolesnikov, D.~Weissenborn, X.~Zhai, T.~Unterthiner, M.~Dehghani, M.~Minderer, G.~Heigold, S.~Gelly \emph{et~al.}, ``An image is worth 16x16 words: Transformers for image recognition at scale,'' \emph{arXiv preprint arXiv:2010.11929}, 2020.

\bibitem{Liu2023}
J.~Liu, H.~Sun, and J.~Katto, ``Learned image compression with mixed transformer-cnn architectures,'' in \emph{2023 IEEE/CVF Conference on Computer Vision and Pattern Recognition (CVPR)}, 2023, pp. 14\,388--14\,397.

\bibitem{TGRSIC_frequency}
S.~Xiang and Q.~Liang, ``Remote sensing image compression based on high-frequency and low-frequency components,'' \emph{IEEE Transactions on Geoscience and Remote Sensing}, vol.~62, pp. 1--15, 2024.

\bibitem{TGRSIC_markov}
Y.~Chong, L.~Zhai, and S.~Pan, ``High-order markov random field as attention network for high-resolution remote-sensing image compression,'' \emph{IEEE Transactions on Geoscience and Remote Sensing}, vol.~60, pp. 1--14, 2022.

\bibitem{TGRSIC_multilevel}
C.~Shi, K.~Shi, F.~Zhu, Z.~Zeng, and L.~Wang, ``A multilevel domain similarity enhancement guided network for remote sensing image compression,'' \emph{IEEE Transactions on Geoscience and Remote Sensing}, vol.~62, pp. 1--19, 2024.

\bibitem{nln}
X.~Wang, R.~Girshick, A.~Gupta, and K.~He, ``Non-local neural networks,'' in \emph{2018 IEEE/CVF Conference on Computer Vision and Pattern Recognition}, 2018, pp. 7794--7803.

\bibitem{autoregressive}
D.~Minnen, J.~Ball{\'{e}}, and G.~Toderici, ``Joint autoregressive and hierarchical priors for learned image compression,'' \emph{CoRR}, vol. abs/1809.02736, 2018.

\bibitem{channel-wise}
D.~Minnen and S.~Singh, ``Channel-wise autoregressive entropy models for learned image compression,'' in \emph{2020 IEEE International Conference on Image Processing (ICIP)}, 2020, pp. 3339--3343.

\bibitem{checkboard}
D.~He, Y.~Zheng, B.~Sun, Y.~Wang, and H.~Qin, ``Checkerboard context model for efficient learned image compression,'' in \emph{2021 IEEE/CVF Conference on Computer Vision and Pattern Recognition (CVPR)}, 2021, pp. 14\,766--14\,775.

\bibitem{ELIC}
D.~He, Z.~Yang, W.~Peng, R.~Ma, H.~Qin, and Y.~Wang, ``Elic: Efficient learned image compression with unevenly grouped space-channel contextual adaptive coding,'' in \emph{2022 IEEE/CVF Conference on Computer Vision and Pattern Recognition (CVPR)}, 2022, pp. 5708--5717.

\bibitem{CrossNet}
H.~Zheng, M.~Ji, H.~Wang, Y.~Liu, and L.~Fang, ``{CrossNet:} {An} end-to-end reference-based super resolution network using cross-scale warping,'' in \emph{Proceedings of the European conference on computer vision (ECCV)}, 2018, pp. 88--104.

\bibitem{SRNTT}
Z.~Zhang, Z.~Wang, Z.~Lin, and H.~Qi, ``Image super-resolution by neural texture transfer,'' in \emph{2019 IEEE/CVF Conference on Computer Vision and Pattern Recognition (CVPR)}, 2019, pp. 7974--7983.

\bibitem{AllYouNeed}
A.~Vaswani, N.~Shazeer, N.~Parmar, J.~Uszkoreit, L.~Jones, A.~N. Gomez, L.~Kaiser, and I.~Polosukhin, ``Attention is all you need,'' in \emph{Advances in Neural Information Processing Systems}, 2017, pp. 5998--6008.

\bibitem{TTSR}
F.~Yang, H.~Yang, J.~Fu, H.~Lu, and B.~Guo, ``Learning texture transformer network for image super-resolution,'' in \emph{2020 IEEE/CVF Conference on Computer Vision and Pattern Recognition (CVPR)}, 2020, pp. 5790--5799.

\bibitem{MASA}
L.~Lu, W.~Li, X.~Tao, J.~Lu, and J.~Jia, ``Masa-sr: Matching acceleration and spatial adaptation for reference-based image super-resolution,'' in \emph{2021 IEEE/CVF Conference on Computer Vision and Pattern Recognition (CVPR)}, 2021, pp. 6364--6373.

\bibitem{li2023rfd}
M.~Li, L.~Shen, P.~Ye, G.~Feng, and Z.~Wang, ``Rfd-ecnet: Extreme underwater image compression with reference to feature dictionary,'' in \emph{Proceedings of the IEEE/CVF International Conference on Computer Vision}, 2023, pp. 12\,980--12\,989.

\bibitem{underwater}
M.~Li, L.~Shen, X.~Hua, and Z.~Tian, ``Euicn: An efficient underwater image compression network,'' \emph{IEEE Transactions on Circuits and Systems for Video Technology}, vol.~34, no.~7, pp. 6474--6488, 2024.

\bibitem{luo2022memory}
A.~Luo, H.~Sun, J.~Liu, and J.~Katto, ``Memory-efficient learned image compression with pruned hyperprior module,'' in \emph{2022 IEEE International Conference on Image Processing (ICIP)}.\hskip 1em plus 0.5em minus 0.4em\relax IEEE, 2022, pp. 3061--3065.

\bibitem{luo2023pts}
A.~Luo, H.~Sun, J.~Liu, F.~Lin, and J.~Katto, ``Pts-lic: Pruning threshold searching for lightweight learned image compression,'' in \emph{2023 IEEE International Conference on Visual Communications and Image Processing (VCIP)}.\hskip 1em plus 0.5em minus 0.4em\relax IEEE, 2023, pp. 1--5.

\bibitem{fang2023fully}
Y.~Fang, W.~Fei, S.~Li, W.~Dai, C.~Li, J.~Zou, and H.~Xiong, ``Fully integerized end-to-end learned image compression,'' in \emph{2023 Data Compression Conference (DCC)}.\hskip 1em plus 0.5em minus 0.4em\relax IEEE, 2023, pp. 337--337.

\bibitem{he2023efficient}
Z.~He, L.~Luo, L.~Zhang, H.~Guo, and C.~Zhu, ``Efficient lightweight attention based learned image compression,'' in \emph{2023 IEEE International Conference on Visual Communications and Image Processing (VCIP)}.\hskip 1em plus 0.5em minus 0.4em\relax IEEE, 2023, pp. 1--5.

\bibitem{fu2023asymmetric}
H.~Fu, F.~Liang, J.~Liang, B.~Li, G.~Zhang, and J.~Han, ``Asymmetric learned image compression with multi-scale residual block, importance scaling, and post-quantization filtering,'' \emph{IEEE Transactions on Circuits and Systems for Video Technology}, vol.~33, no.~8, pp. 4309--4321, 2023.

\bibitem{wang2024asymllic}
S.~Wang, Z.~Cheng, D.~Feng, G.~Lu, L.~Song, and W.~Zhang, ``Asymllic: Asymmetric lightweight learned image compression,'' in \emph{2024 IEEE International Conference on Visual Communications and Image Processing (VCIP)}.\hskip 1em plus 0.5em minus 0.4em\relax IEEE, 2024, pp. 1--5.

\bibitem{bao2025shiftlic}
Y.~Bao, W.~Tan, C.~Jia, M.~Li, Y.~Liang, and Y.~Tian, ``Shiftlic: Lightweight learned image compression with spatial-channel shift operations,'' \emph{IEEE Transactions on Circuits and Systems for Video Technology}, 2025.

\bibitem{div2k}
E.~Agustsson and R.~Timofte, ``Ntire 2017 challenge on single image super-resolution: Dataset and study,'' in \emph{2017 IEEE Conference on Computer Vision and Pattern Recognition Workshops (CVPRW)}, 2017, pp. 1122--1131.

\bibitem{BM3D}
K.~Dabov, A.~Foi, V.~Katkovnik, and K.~Egiazarian, ``Image denoising by sparse 3-d transform-domain collaborative filtering,'' \emph{IEEE Transactions on Image Processing}, vol.~16, no.~8, pp. 2080--2095, 2007.

\bibitem{GLCM}
R.~M. Haralick, K.~Shanmugam, and I.~Dinstein, ``Textural features for image classification,'' \emph{IEEE Transactions on Systems, Man, and Cybernetics}, vol. SMC-3, no.~6, pp. 610--621, 1973.

\bibitem{VGG}
K.~Simonyan and A.~Zisserman, ``Very deep convolutional networks for large-scale image recognition,'' \emph{Computer Science}, 2014.

\bibitem{minnen}
D.~Minnen, J.~Ball{\'e}, and G.~D. Toderici, ``Joint autoregressive and hierarchical priors for learned image compression,'' \emph{Advances in neural information processing systems}, vol.~31, 2018.

\bibitem{compressai}
J.~Bégaint, F.~Racapé, S.~Feltman, and A.~Pushparaja, ``Compressai: a pytorch library and evaluation platform for end-to-end compression research,'' Nov. 2020.

\bibitem{Adam}
D.~P. Kingma and J.~Ba, ``Adam: A method for stochastic optimization,'' \emph{arXiv preprint arXiv:1412.6980}, 2014.

\bibitem{ms-ssim}
Z.~Wang, E.~Simoncelli, and A.~Bovik, ``Multiscale structural similarity for image quality assessment,'' in \emph{The Thrity-Seventh Asilomar Conference on Signals, Systems \& Computers, 2003}, vol.~2, 2003, pp. 1398--1402 Vol.2.

\bibitem{lpips}
R.~Zhang, P.~Isola, A.~A. Efros, E.~Shechtman, and O.~Wang, ``The unreasonable effectiveness of deep features as a perceptual metric,'' in \emph{2018 IEEE/CVF Conference on Computer Vision and Pattern Recognition}, 2018, pp. 586--595.

\bibitem{bdpsnr}
G.~Bjontegaard, ``Calculation of average psnr differences between rd-curves,'' \emph{ITU SG16 Doc. VCEG-M33}, 2001.

\bibitem{fid}
M.~Heusel, H.~Ramsauer, T.~Unterthiner, B.~Nessler, and S.~Hochreiter, ``Gans trained by a two time-scale update rule converge to a local nash equilibrium,'' \emph{Advances in neural information processing systems}, vol.~30, 2017.

\bibitem{jpeg}
G.~K. Wallace, ``The jpeg still picture compression standard,'' \emph{Communications of the ACM}, vol.~34, no.~4, pp. 30--44, 1991.

\end{thebibliography}
\bibliographystyle{IEEEtran}
\vspace{-0.5cm}
\begin{IEEEbiography}[{\includegraphics[width=1in,height=1.25in,clip,keepaspectratio] {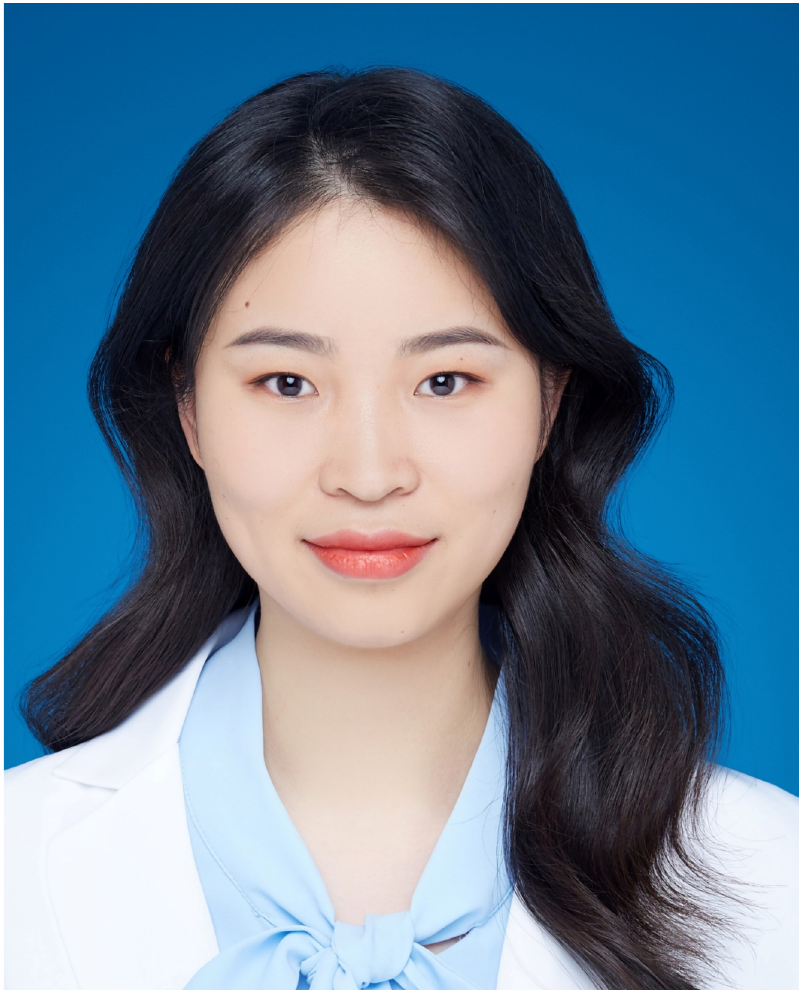}}]{Qing Ding} received the B.S. and M.S. degrees from the School of Communication and Information Engineering, Shanghai University, Shanghai, China, in 2018 and 2021. She is currently pursuing the Ph.D. degree from the School of Electronic and Information Engineering, Beihang University, Beijing, China. Her research interests include video coding, deep learning, video enhancement, and learned image compression.
\end{IEEEbiography}
\vspace{-0.5cm}
\begin{IEEEbiography}[{\includegraphics[width=1in,height=1.25in,clip,keepaspectratio] {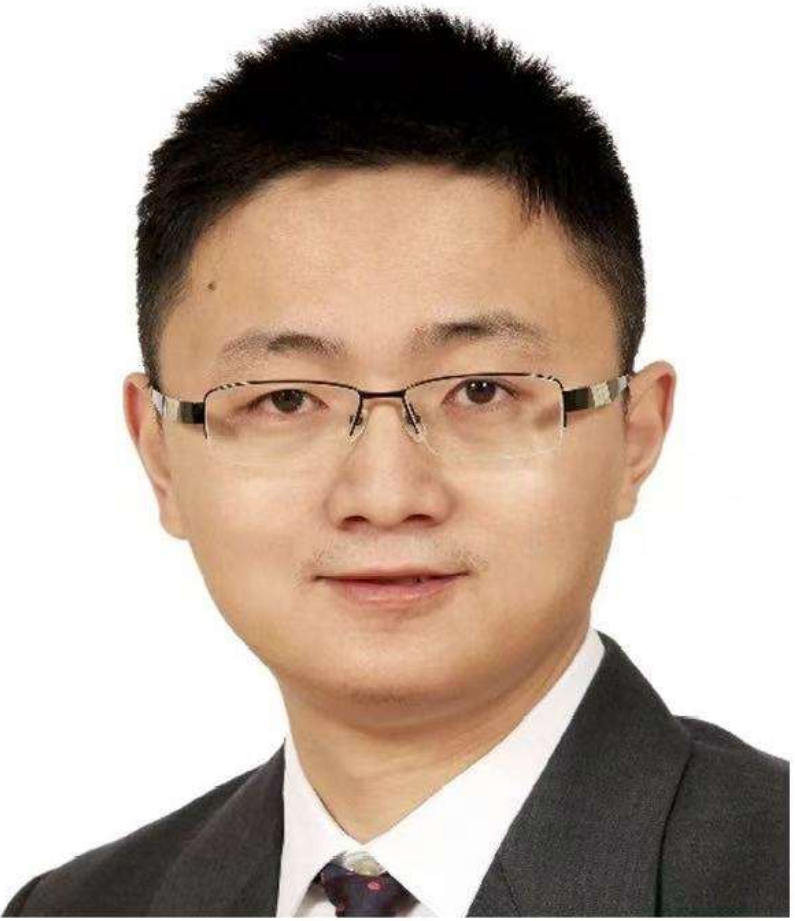}}]{Mai Xu} (Senior Member, IEEE) received the B.S. degree from Beihang University, Beijing, China, in 2003, the M.S. degree from Tsinghua University, Beijing, in 2006, and the Ph.D. degree from Imperial College London, London, U.K., in 2010. From 2010 to 2012, he was a Research Fellow with the Department of Electrical Engineering, Tsinghua University. Since 2013, he has been with Beihang University, where he was an Associate Professor and promoted to Full Professor in 2019. From 2014 to 2015, he was a Visiting Researcher of MSRA. He has authored or coauthored more than 200 technical papers in international journals and conference proceedings, such as International Journal of Computer Vision, IEEE Transactions on Pattern Analysis and Machine Intelligence, IEEE Transactions on Image Processing, IEEE Journal of Selected Topics in Signal Processing, Conference on Computer Vision and Pattern Recognition, International Conference on Computer Vision, European Conference on Computer Vision, and AAAI. His main research interests include image processing and computer vision. He was the recipient of best/top paper awards of IEEE/ACM conferences, such as ACM MM. He was an Associate Editor for IEEE Transactions on Image Processing and IEEE Transactions on Multimedia, the Lead Guest Editor of IEEE Journal of Selected Topics in Signal Processing, and the Area Chair or a TPC Member of many conferences, such as ICME and AAAI. He is an elected Member of Multimedia Signal Processing Technical Committee and IEEE Signal Processing Society.
\end{IEEEbiography}
\vspace{-0.5cm}
\begin{IEEEbiography}[{\includegraphics[width=1in,height=1.25in,clip,keepaspectratio] {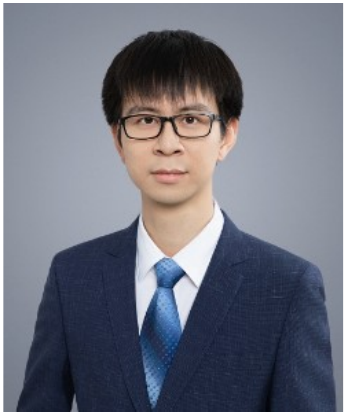}}]{Shengxi Li} (Member, IEEE) received the Ph.D. degree in electrical and electronic engineering from Imperial College London, London, U.K., in 2021. He is currently a Professor with the School of Electronic and Information Engineering, Beihang University, Beijing, China. His research interests include generative models, statistical signal processing, and machine learning. He was a recipient of the Young Investigator Award of International Neural Network Society.
\end{IEEEbiography}
\vspace{-0.5cm}
\begin{IEEEbiography}[{\includegraphics[width=1in,height=1.25in,clip,keepaspectratio] {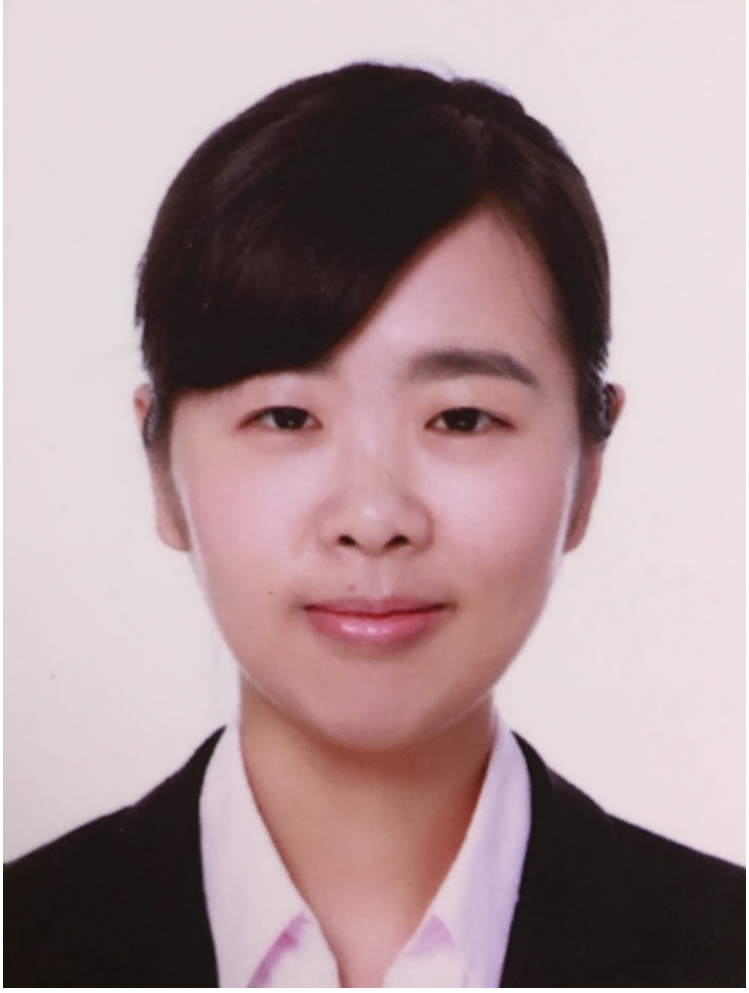}}]{Xin Deng} (Member, IEEE) received a master's degree in electrical information engineering from Beihang University, Beijing, China, in 2016, and a Ph.D. degree in electrical and electronic engineering from Imperial College London, London, U.K., in March 2020. She is currently an associate professor in the Department of Cyber Science and Technology, Beihang University, Beijing, China. Her research interests include sparse coding with applications in image and video processing, machine learning, and multimodal signal processing.
\end{IEEEbiography}
\vspace{-0.5cm}
\begin{IEEEbiography}[{\includegraphics[width=1in,height=1.25in,clip,keepaspectratio] {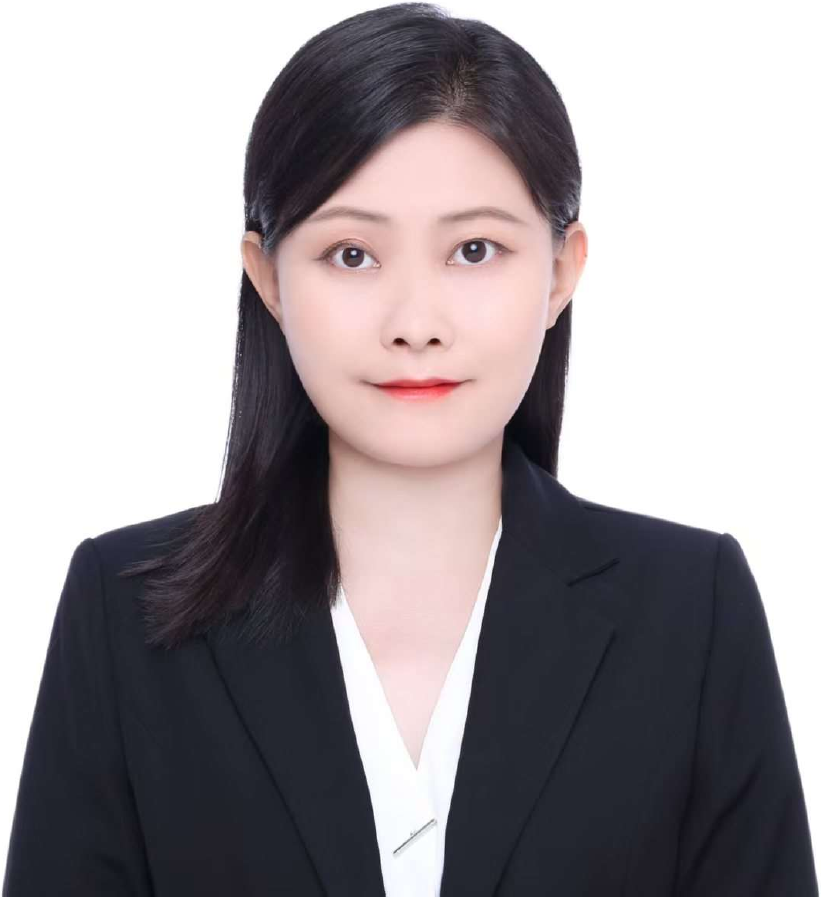}}]{Xin Zou} received the Master's degree from Changchun University of Science and Technology in 2009. Since then, she has been working at Beijing Institute of Spacecraft System Engineering. Her research interests include spacecraft system design and image processing.
\end{IEEEbiography}
\end{document}